\documentclass[preprint,12pt]{elsarticle}

\usepackage{soul, color}
\usepackage{graphicx}
\usepackage{adjustbox}
\usepackage{booktabs,makecell}
\usepackage{makecell}
\usepackage{tabularx}
\usepackage{threeparttable, tablefootnote}
\usepackage{amsmath}
\newcommand{\linenumbers}{}

\usepackage{newunicodechar}

\usepackage[colorlinks,allcolors=black]{hyperref}
\usepackage{hyperref}
\hypersetup{colorlinks,allcolors=black}

\usepackage{graphicx}
\usepackage{multirow}
\usepackage[dvipsnames]{xcolor}
\usepackage{soul}
\usepackage{color}
\usepackage{url}
\usepackage{float}
\usepackage{boldline}
\usepackage{xcolor,colortbl}
\usepackage{tikz}

\usepackage{pifont}

\usepackage{multirow}
\usepackage{graphicx}
\usepackage{amssymb}
\usepackage{soul}
\usepackage[normalem]{ulem}
\usepackage{xcolor}

\usepackage{xspace}
\usepackage{tabularx}

\journal{}

\begin{document}

\begin{frontmatter}


\title{Hierarchical Deep Feature Fusion and Ensemble Learning for Enhanced Brain Tumor MRI Classification}

\author{Zahid Ullah$^{1}$, Jihie Kim$^{1,*}$}

\address{%
$^{1}$ \quad Department of Computer Science and Artificial Intelligence, Dongguk University, Seoul 04620, Republic of Korea \\ } 

\begin{abstract}
Accurate brain tumor classification is crucial in medical imaging to ensure reliable diagnosis and effective treatment planning. This study introduces a novel double ensembling framework that synergistically combines pre-trained deep learning (DL) models for feature extraction with optimized machine learning (ML) classifiers for robust classification. The framework incorporates comprehensive preprocessing and data augmentation of brain magnetic resonance images (MRI), followed by deep feature extraction using transfer learning with pre-trained Vision Transformer (ViT) networks. The novelty lies in the dual-level ensembling strategy: feature-level ensembling, which integrates deep features from the top-performing ViT models, and classifier-level ensembling, which aggregates predictions from hyperparameter-optimized ML classifiers. Experiments on two public Kaggle MRI brain tumor datasets demonstrate that this approach significantly surpasses state-of-the-art methods, underscoring the importance of feature and classifier fusion. The proposed methodology also highlights the critical roles of hyperparameter optimization (HPO) and advanced preprocessing techniques in improving diagnostic accuracy and reliability, advancing the integration of DL and ML for clinically relevant medical image analysis.

\end{abstract}

\end{frontmatter}

\linenumbers

\section{Introduction}
\label{intro}
Magnetic Resonance Imaging (MRI) is a cornerstone of modern medical diagnostics, known for its exceptional ability to non-invasively visualize complex anatomical structures and pathological conditions with high-resolution, multi-planar imaging \cite{amin2023brain}. This capability makes MRI invaluable for detecting and characterizing a wide range of diseases, particularly brain tumors \cite{xu2024cross}. Its sensitivity to subtle variations in tissue composition and physiological processes allows for precise differentiation between normal and abnormal tissues. By leveraging the unique magnetic properties of various tissue types, MRI provides detailed insights into the size, shape, and location of tumors, facilitating accurate diagnosis and treatment planning \cite{t2024enhancing}. Beyond initial detection, MRI plays a critical role in monitoring treatment responses and identifying tumor recurrence, directly influencing patient outcomes \cite{zulfiqar2023multi}. Despite these strengths, the interpretation of MRI images often relies on the expertise of human specialists, which introduces subjectivity and potential for error \cite{aygun2018multi}. This dependence underscores the need for advanced analytical tools to enhance diagnostic precision and consistency, addressing the challenges of human variability in image interpretation.

The manual evaluation of MRIs is fraught with limitations, including inter-observer variability, fatigue-induced errors, and difficulties in distinguishing subtle differences between tumor and normal tissues \cite{dehkordi2022brain}. These challenges are particularly pronounced when tumors are small, poorly defined, or located in complex anatomical regions, leading to inconsistencies in diagnosis \cite{tjahyaningtijas2018brain}. Additionally, the vast volume of MRI data generated in clinical practice places a considerable burden on radiologists, increasing the risk of overlooking critical findings. To address these challenges and improve the accuracy and efficiency of brain tumor diagnosis, automated image analysis techniques are increasingly being developed to complement human expertise. These methods, powered by machine learning (ML) and deep learning (DL) algorithms, offer the potential to standardize image interpretation, minimize inter-observer variability, and detect subtle tumor characteristics that might elude human observers \cite{madgi2021brain}. The emergence of artificial intelligence in medical imaging marks a transformative shift, revolutionizing the detection, diagnosis, and management of brain tumors with enhanced precision and reliability \cite{gunasekara2020feasibility}.

Accurate and reliable classification of brain tumors is essential for effective treatment planning and improved patient outcomes, yet it remains a significant challenge in medical image analysis \cite{dehkordi2022brain}, \cite{gundogan2025novel}. This complexity stems from the substantial heterogeneity in tumor morphology, texture, and contrast, which varies not only between patients but also within different regions of the same tumor \cite{abd2019review}. Traditional ML techniques, which rely heavily on handcrafted features, often fail to capture this complexity, limiting their generalizability and robustness \cite{t2024enhancing}. In contrast, DL methodologies, particularly convolutional neural networks (CNNs), have demonstrated remarkable success in automatically extracting intricate features directly from medical images, achieving state-of-the-art (SOTA) performance in brain tumor classification \cite{magadza2021deep}. However, DL models typically require large, accurately labeled datasets for effective training—a challenge in the medical domain due to the time-intensive and costly nature of expert annotation. Brain tumors significantly impact patients' lifespan and quality of life, making early and accurate diagnosis critical. ML and image processing techniques have the potential to automate diagnostic workflows, improving both accuracy and reliability \cite{madgi2021brain}. Furthermore, DL has revolutionized healthcare by enhancing recognition, prediction, and diagnosis across various medical domains, including brain tumors, lung cancer, cardiovascular conditions, and retinal diseases \cite{nadeem2020brain}. This progress underscores the transformative potential of advanced AI methodologies in addressing complex medical imaging challenges.

Classifying brain tumors from MRIs is a challenging task due to the heterogeneity of tumor types and subtle variations in their imaging characteristics. The ML models, such as Support Vector Machines (SVM), Random Forests (RF), and deep neural networks (DNNs), have been employed for this purpose, achieving varying levels of success depending on the quality of extracted features and the classification algorithm used \cite{faradibah2023comparison, latif2022glioma, ahmad2022performance}. Pre-training techniques are particularly valuable for enhancing the performance of Vision Transformers (ViTs), especially when working with limited medical image datasets \cite{takahashi2024comparison}. Transfer learning, which involves fine-tuning a model pre-trained on a large dataset for a specific task, has been shown to improve generalization and reduce overfitting in medical imaging applications \cite{matsoukas2023pretrained}. Recent research on ViTs for brain tumor classification has reported promising results, highlighting their ability to capture long-range dependencies and extract meaningful features with high accuracy and robustness \cite{dahan2022surface}. These attributes make ViTs well-suited for addressing the complexities of brain tumor classification.

To overcome the limitations of traditional ML and purely DL approaches, a new hybrid methodology is proposed that combines the strengths of both paradigms. Pre-trained ViTs models, trained on large datasets, are utilized as robust feature extractors to capture salient and discriminative attributes from brain MRI scans. ViTs offer a compelling alternative to CNNs for feature extraction, particularly in medical imaging \cite{feng2022multimodal}. By leveraging self-attention mechanisms, ViTs effectively capture global relationships within an image, identifying intricate dependencies that may be overlooked by traditional convolutional methods \cite{takahashi2024comparison}. Unlike CNNs, which rely on localized receptive fields, ViTs process images as sequences of patches, transforming image analysis into a sequence-to-sequence learning problem \cite{dahan2024multiscale}. This approach enables the model to learn contextual dependencies across the entire image, fostering a holistic understanding of inter-regional relationships.

ViTs have demonstrated promising results in various computer vision tasks and are particularly valuable in medical image analysis, especially when labeled data is limited. The features extracted by these pre-trained models are subsequently fed into ML classifiers, such as SVM or RF, to distinguish between tumorous and normal MRI scans. This hybrid strategy capitalizes on the representational power of pre-trained models while maintaining the flexibility and efficiency of ML classifiers \cite{thakur2024deep}.

The integration of DL and ML methodologies has become increasingly vital in biomedical sciences, providing effective tools for analyzing high-dimensional and multimodal data \cite{babayomi2023convolutional}. These techniques have been extensively applied in brain image analysis, aiding in the development of diagnostic and classification systems for conditions such as strokes, psychiatric disorders, epilepsy, neurodegenerative diseases, and demyelinating disorders \cite{zhu2019applications}. Furthermore, ML and computer vision techniques have the potential to revolutionize radiology workflows by automating the prioritization of imaging studies, such as those involving suspected intracranial hemorrhage, thereby enhancing diagnostic efficiency and accuracy.


To harness the complementary strengths of diverse pre-trained DL architectures, we propose a novel feature ensemble approach for brain tumor classification. Our methodology includes multiple variations: a base version, versions incorporating normalization and principal component analysis ((PCA) i.e., PCA has been used to reduce the dimensionality of the extracted deep features, retaining the most informative components), versions using the synthetic minority over-sampling technique ((SMOTE) i.e., to addresses class imbalance in the datasets by generating synthetic samples for underrepresented tumor classes), and combinations of normalization, PCA, and SMOTE. The core of this approach lies in an innovative feature evaluation and selection mechanism, where deep features extracted from thirteen pre-trained DL models are rigorously assessed using nine distinct ML classifiers. A custom-designed criterion ensures the retention of only the most salient and discriminative features for subsequent analysis. The selected features are combined into a synthetic feature vector by concatenating outputs from the top-performing two or three feature extractor models. This fusion enables the integration of complementary information from multiple DL architectures, resulting in a more robust and discriminative feature representation compared to relying on individual models. These enhanced feature vectors are then input into various ML classifiers to yield the final classification prediction. By integrating features from diverse DL models, our approach surpasses conventional methods that rely solely on traditional feature extraction techniques. This innovative ensemble strategy not only improves classification performance but also establishes a more comprehensive and effective framework for brain tumor diagnosis.

Specifically, our approach employs a double ensemble strategy to enhance brain tumor classification accuracy and robustness. First, we combine features extracted from the top-2 and top-3 performing DL models into aggregated feature representations. These combined features are then input into nine distinct ML classifiers for binary classification. Additionally, features extracted individually from the top-5 pre-trained DL models are presented to an ensemble of the top-2 or top-3 performing ML classifiers to generate the final prediction.

This ensemble strategy highlights the significance of combining multiple models to improve predictive accuracy and generalization. By amalgamating outputs from diverse models, the approach effectively mitigates individual biases and variance, resulting in enhanced performance metrics and more reliable classifications. The development of this robust framework addresses key challenges, including dataset variability, preprocessing requirements, and architectural complexities, establishing a scalable and effective system for brain tumor diagnosis.

In summary, the major contribution of this study are listed as follows:

\begin{itemize}
    \item We introduced a hybrid approach combining feature-level ensembling of pre-trained ViTs models with classifier-level ensembling of fine-tuned ML classifiers, significantly enhancing brain tumor classification accuracy.
    \item We implemented extensive preprocessing techniques and data augmentation strategies to improve data quality and address challenges such as noise and variability in MRI datasets.
    \item We conducted systematic hyperparameter tuning for multiple ML classifiers, demonstrating their pivotal role in achieving superior performance and diagnostic reliability.
    \item We validated the proposed framework on two publicly available Kaggle MRI brain tumor datasets, achieving SOTA performance and showcasing the effectiveness of integrating DL and ML models for medical image analysis.
\end{itemize}

The code implementation of our approach will be openly shared after publication to promote reproducibility and facilitate further research. Code Link: \url{https://github.com/Zahid672/ViT_Ensembling_Brain_Tumor_Classification}

The rest of the paper is structured as follows: Section \ref{related}
reviews the related work. Section \ref{pm} introduces the proposed methodology.
Section \ref{experimental} shows and analyzes the experimental setup. Finally, Section \ref{results} and \ref{discussion} discuss the results and discussion, respectively, and point out future directions, and Section \ref{con} concludes this paper.

\section{Related Work}\label{related}
DL methodologies have revolutionized numerous healthcare domains, demonstrating unparalleled efficacy in recognition, prediction, and diagnosis within areas such as pathology, brain tumor analysis, lung cancer detection, abdominal imaging, cardiac assessments, and retinal evaluations \cite{nadeem2020brain}. The complexity and scale of healthcare services highlight the vital importance of computer-aided diagnostic techniques, especially for the timely and precise detection of brain tumors using MRI scans. We have categorized the related work into two main groups: classical machine learning-based techniques and deep learning-based techniques, as summarized in Table \ref{relatedwork}.

\subsection{Brain Tumor Classification using Traditional Machine Learning}
The integration of ML methodologies into biomedical sciences has facilitated the analysis of complex, high-dimensional data, thereby offering avenues for enhanced disease detection and treatment \cite{babayomi2023convolutional}. For instance, Ural et al. \cite{ural2018computer} presents a computer-based approach for brain tumor detection using probabilistic neural network (PNN). The method employs k-means clustering for image segmentation and fuzzy c-means for feature extraction, enabling effective tumor identification. The approach was tested on 25 MRI samples, achieving an accuracy of 90\%. While the method demonstrates promising results, its primary limitation lies in the small dataset size, which limits generalizability and robustness across diverse clinical scenarios. Additionally, the reliance on handcrafted features and conventional clustering techniques may struggle with complex, high-dimensional data, highlighting the need for more scalable and automated DL solutions in this domain. 

\begin{table}[!ht]
\centering
\caption{Comparison of related studies for brain tumor classification.}
\scalebox{0.7}{
\begin{tabular}{ccccccc}
\hline
\textbf{Research} & \textbf{Solution}                                                                                              & \textbf{AI Models}                                                                      & \textbf{Objective}                                                                                    & \textbf{Data Size}                                          & \textbf{\begin{tabular}[c]{@{}c@{}}Feature   Extraction\\  Technique\end{tabular}} & \textbf{Accuracy} \\ \hline
 \cite{ural2018computer}          & \multirow{5}{*}{\begin{tabular}[c]{@{}c@{}}Classical   \\ Machine\\  Learning-\\ based solutions\end{tabular}} & PNN                                                                                     & \begin{tabular}[c]{@{}c@{}}Brain tumor  \\  detection\end{tabular}                                    & \begin{tabular}[c]{@{}c@{}}25 MRI \\ samples\end{tabular}   & \begin{tabular}[c]{@{}c@{}}k-mean with  \\ fuzzy c-mean\end{tabular}               & 90\%              \\
 \cite{ullah2020hybrid}          &                                                                                                                & \begin{tabular}[c]{@{}c@{}}Feed-forward   \\ neural network\end{tabular}                & \begin{tabular}[c]{@{}c@{}}Brain MRI   \\ classification \\ into normal \\ and abnormal\end{tabular}  & \begin{tabular}[c]{@{}c@{}}71 MRI \\ samples\end{tabular}   & DWT                                                                                & 95.8\%            \\
\cite{varuna2018identification}          &                                                                                                                & \begin{tabular}[c]{@{}c@{}}Probabilistic   \\ neural network\end{tabular}               & \begin{tabular}[c]{@{}c@{}}Brain MRI   \\ classification \\ into  normal \\ and abnormal\end{tabular} & \begin{tabular}[c]{@{}c@{}}650 MR\\ samples\end{tabular}    & GLCM                                                                               & 95.0\%            \\
\cite{kharrat2010hybrid}          &                                                                                                                & \begin{tabular}[c]{@{}c@{}}Hybrid   method\\ Genetic algorithm\\  with SVM\end{tabular} & \begin{tabular}[c]{@{}c@{}}Brain MRI   \\ classification\\  into normal \\ and abnormal\end{tabular}  & \begin{tabular}[c]{@{}c@{}}83 MRI \\ samples\end{tabular}   & \begin{tabular}[c]{@{}c@{}}Wavelet-based  \\  features\end{tabular}                & 98.14\%           \\
\cite{rajan2019brain}          &                                                                                                                & SVM                                                                                     & Tumor   detection                                                                                     & \begin{tabular}[c]{@{}c@{}}41 MRI \\ samples\end{tabular}   & Adaptive GLCM                                                                      & 98.0\%            \\ \hline
\cite{ccinar2020detection}          & \multirow{7}{*}{\begin{tabular}[c]{@{}c@{}}Deep   \\ Learning-\\ based \\ solutions\end{tabular}}              & CNN models                                                                              & \begin{tabular}[c]{@{}c@{}}Brain tumor \\ detection and \\ classification\end{tabular}                & \begin{tabular}[c]{@{}c@{}}253 MRI \\ samples\end{tabular}  & CNN                                                                                & 97.2\%            \\
 \cite{mehnatkesh2023intelligent}           &                                                                                                                & ResNet                                                                                  & \begin{tabular}[c]{@{}c@{}}Brain tumor \\ classification\end{tabular}                                 & \begin{tabular}[c]{@{}c@{}}3064 MRI\\  samples\end{tabular} & CNN                                                                                & 98.69\%           \\
\cite{deepak2019brain}          &                                                                                                                & Transfer   learning                                                                     & \begin{tabular}[c]{@{}c@{}}Brain tumor \\ classification\end{tabular}                                 & \begin{tabular}[c]{@{}c@{}}3064 MRI\\ samples\end{tabular}  & GoogleNet                                                                          & 98.0\%            \\
\cite{diaz2021deep}          &                                                                                                                & CNN                                                                                     & \begin{tabular}[c]{@{}c@{}}Brain tumor\\   classification\end{tabular}                                & \begin{tabular}[c]{@{}c@{}}3064 MRI\\ samples\end{tabular}  & CNN                                                                                & 97.3\%            \\
\cite{khan2022accurate}           &                                                                                                                & CNN                                                                                     & \begin{tabular}[c]{@{}c@{}}Brain tumor\\  detection\end{tabular}                                      & \begin{tabular}[c]{@{}c@{}}3064 MRI \\ samples\end{tabular} & CNN                                                                                & 97.8\%            \\
\cite{paul2017deep}         &                                                                                                                & CNN                                                                                     & \begin{tabular}[c]{@{}c@{}}Brain tumor  \\ classification\end{tabular}                                & \begin{tabular}[c]{@{}c@{}}3064 MRI \\ samples\end{tabular} & CNN                                                                                & 91.43\%           \\
 \cite{hemanth2018modified}           &                                                                                                                & CNN                                                                                     & \begin{tabular}[c]{@{}c@{}}Brain tumor  \\  classification\end{tabular}                               & \begin{tabular}[c]{@{}c@{}}220 MRI \\ samples\end{tabular}  & CNN                                                                                & 94.5\%            \\ \hline
\end{tabular}
}
\label{relatedwork}
\end{table}

Zahid et al. \cite{ullah2020hybrid} propose a framework combining traditional image preprocessing with DL for classifying brain MRI scans. The method employs median filtering for noise reduction, histogram equalization for contrast enhancement, discrete wavelet transform (DWT) for feature extraction, and color moments (mean, standard deviation, skewness) for dimensionality reduction, followed by a DNN classifier. While achieving 95.8\% accuracy, the approach has key limitations: (1) its performance heavily depends on the sequence of enhancement steps, which may not generalize across diverse MRI datasets with varying noise or contrast profiles; (2) aggressive preprocessing (e.g., histogram equalization) risks distorting subtle pathological features critical for diagnosis; and (3) the evaluation used limited datasets, raising concerns about scalability to real-world clinical settings with heterogeneous imaging protocols. These constraints highlight the need for adaptive enhancement strategies and larger, multi-institutional validation to improve robustness.

Shree and Kumar \cite{varuna2018identification} proposes a methodology combining DWT for feature extraction and a PNN for classification, obtained 95.0\% accuracy on a dataset of 650 MRI images. The use of DWT helps in extracting detailed frequency domain features, making the classification process effective. However, the study's reliance on handcrafted features through DWT limits its adaptability to more complex and varied datasets. Additionally, the PNN classifier, while efficient for smaller datasets, may not scale well or handle the high-dimensional features extracted from larger datasets. The model also lacks robustness against noise and variability in MRI quality, which could impact its practical application in diverse clinical scenarios.

Kharrat et al. \cite{kharrat2010hybrid} proposes a methodology that combines a genetic algorithm (GA) for feature selection with an SVM for classification.  The study reports significant accuracy improvements, demonstrating the effectiveness of GA in optimizing the feature space for SVM classification. However, the primary limitation of this approach lies in its computational intensity, as GA is resource-intensive and time-consuming, especially when dealing with larger datasets. Furthermore, the reliance on handcrafted features limits the model's ability to adapt to varying imaging conditions and complex data structures, highlighting the potential for more automated DL methods to address these challenges.

We observed from the traditional ML techniques for brain tumor classification, such as those relying on handcrafted feature extraction and algorithms like SVM or GA, face several notable limitations. These methods often depend heavily on manual feature engineering, which may not adequately capture the complex and heterogeneous patterns present in brain MRIs, leading to restricted accuracy and poor generalizability across diverse datasets. The high variability in tumor appearance, intensity, and shape can further challenge these models, as they struggle to adapt to the high-dimensional nature of MRI data. Additionally, traditional approaches are sensitive to noise and variations in imaging protocols, and their performance is often constrained by the quality and relevance of the selected features. As a result, these models may underperform when confronted with real-world clinical data, highlighting the need for more automated, robust, and adaptive methods such as DL-based frameworks that can learn hierarchical features directly from raw images. Hence, researchers are increasingly adopting DL methodologies as a cornerstone for advancing brain tumor classification techniques.

\subsection{Brain Tumor Classification using Deep Learning}
For instance, Ahmet and Muhammad \cite{ccinar2020detection} aimed to improve diagnostic accuracy and support computer-aided diagnosis. The authors base their approach on the ResNet50 architecture, modifying it by removing the last five layers and adding eight new layers tailored for tumor classification. Their hybrid model achieves a high accuracy of 97.2\% and is benchmarked against other popular CNN models such as AlexNet, DenseNet201, InceptionV3, and GoogLeNet, outperforming them in their experiments. However, the study has several limitations: the evaluation is conducted on relatively limited datasets, which may not capture the full heterogeneity of real-world clinical images; the model’s generalizability to unseen data and different MRI protocols is not thoroughly assessed; and the approach focuses primarily on classification without addressing tumor localization or segmentation, which are critical for clinical applications. These factors suggest the need for more diverse datasets and integration with localization methods to enhance clinical utility.

Hossein et al. \cite{mehnatkesh2023intelligent} introduce a deep residual learning framework for categorizing brain tumors, leveraging an optimization-driven ResNet architecture enhanced by evolutionary algorithms such as ant colony optimization and differential evolution. This approach automates the design and hyperparameter tuning of the deep residual network, aiming to improve classification accuracy and reduce manual intervention. The proposed framework demonstrates strong performance, achieving an average accuracy of approximately 98.7\% on benchmark datasets, and highlights its potential for robust and scalable tumor categorization in clinical imaging workflows. However, the method has notable limitations: its reliance on evolutionary optimization increases computational complexity and resource requirements, potentially hindering real-time or large-scale clinical deployment.


Deepak et al. \cite{deepak2019brain} present a brain tumor classification system that leverages deep CNN features via transfer learning to distinguish among brain tumors. The authors utilize pre-trained GoogLeNet to extract features from MRIs, followed by the integration of proven classifier models to perform the final classification. Their approach employs a patient-level five-fold cross-validation on a publicly available dataset and achieves a mean classification accuracy of 98\%. The study further demonstrates that transfer learning is particularly effective when training data is limited, and provides analytical insights into misclassification cases. Despite its strengths, the study has notable limitations. The research does not explore ensemble strategies or fine-tuning of the pre-trained models, potentially limiting the performance. The lack of detailed comparisons with traditional ML methods or other transfer learning frameworks also restricts insights into the method's relative advantages.

Francisco et al. \cite{diaz2021deep} presents a fully automated system that leverages a multiscale deep CNN to brain tumors from MRI. The proposed model processes 2D MRI slices using a sliding window approach and employs three parallel pathways with different convolutional kernel sizes to capture image features at multiple spatial resolutions, mimicking the human visual system. This architecture enables the model to distinguish between normal and abnormal MRIs. However, the approach has its limitations. The computational demands of multiscale CNNs are significantly higher, making real-time application or deployment on resource-constrained systems challenging. Additionally, the study does not thoroughly address the role of preprocessing and data augmentation, which could affect performance on noisy or imbalanced datasets. Furthermore, the evaluation primarily focuses on classification and segmentation accuracy, without delving into robustness against diverse imaging conditions or comparisons with SOTA ensemble or hybrid models.

Khan et al. \cite{khan2022accurate} proposes two deep CNN models for accurate brain tumor detection and classification using MRIs. The proposed framework leverages advanced CNN architectures to automatically extract discriminative features, achieving high classification accuracy. By adopting transfer learning, the model benefits from pre-trained weights on extensive datasets, allowing it to perform effectively even on smaller, domain-specific datasets. The study demonstrates promising results in terms of classification precision and computational efficiency, showcasing the potential of CNNs in clinical applications. However, the work has certain limitations. The reliance on a single type of CNN architecture without exploring ensemble or hybrid approaches may restrict the model's robustness to varied data distributions. Additionally, the study does not address issues like imbalanced datasets or noisy MRI scans, which can significantly impact performance in real-world scenarios.  Furthermore, the lack of comparative analysis with other SOTA methods limits the broader applicability and validation of the proposed approach.

Paul et al. \cite{paul2017deep} investigates the potential of DL techniques for classifying brain tumors using MRI scans. The study utilizes CNNs to automatically learn hierarchical features from MRI datasets, aiming to improve classification accuracy over traditional ML approaches. The authors demonstrate the ability of CNNs to achieve significant performance gains in tumor classification tasks, particularly through the use of architectures designed to handle high-dimensional medical imaging data. This work underscores the transformative potential of DL in medical imaging applications. However, the study has some limitations. The dataset used for training and evaluation is not explicitly detailed, raising questions about its diversity and size, which are critical for generalizability. The study also does not consider the impact of imbalanced datasets or preprocessing steps, such as augmentation, which are often required to enhance model robustness. Finally, the lack of a comparative analysis with other contemporary methods limits the assessment of the proposed approach's effectiveness relative to existing techniques.

Recent research has focused on employing DL models, such as CNNs and ViTs, to automate the analysis of brain MRIs, achieving promising results in terms of accuracy and robustness \cite{dehkordi2022brain}.

The synergy between the long-range dependencies captured by transformers and the local representations learned by CNNs has led to advanced architectures that have improved performance in various medical image analysis tasks \cite{shen2023movit}. The efficacy of ViTs in feature extraction stems from their ability to model the correlation between image patches, which is particularly relevant for capturing subtle variations in tumor morphology and texture \cite{feng2022multimodal}. ViTs represent a paradigm shift in image analysis, diverging from traditional CNN-based approaches by leveraging self-attention mechanisms to capture long-range dependencies within images \cite{aygun2018multi}. Transformers have demonstrated remarkable success in various computer vision tasks, including image classification, object detection, and semantic segmentation, owing to their ability to model contextual relationships effectively \cite{xia2023recent}. The foundational architecture of a ViTs involves partitioning an input image into a sequence of patches, which are then linearly embedded into a high-dimensional space. These embedded patches are processed through a series of transformer layers, each comprising a multi-head self-attention mechanism and a feed-forward network, facilitating the learning of intricate relationships between image regions.  Within the domain of medical image analysis, ViTs have garnered increasing attention due to their capacity to handle the inherent complexity and variability of medical images, offering potential advantages over conventional CNNs. The applicability of transformers in computer vision has prompted researchers to explore their utility in medical imaging, particularly for tasks such as classification, segmentation, registration, and reconstruction, achieving SOTA performance on standard medical datasets \cite{henry2022vision}.

\subsection{Summary}
 In summary, the majority of existing research indicates that DL techniques achieve significantly higher accuracy in brain MRI classification compared to traditional ML methods. However, DL systems require substantial amounts of training data to outperform conventional ML approaches. Recent studies have established DL as a leading method in medical image analysis, though these techniques also present specific challenges that must be addressed in brain tumor classification and segmentation tasks. In this study, we propose a fully automated hybrid methodology for brain tumor classification, integrating two main components: pre-trained DL models for extracting deep features from brain MRI and ML classifiers for precise brain MRI classification. Our approach is distinguished by four unique steps. 
First, deep features are extracted from MRIs using pre-trained DL models. Second, these features are rigorously analyzed using nine different ML classifiers. Third, the top-performing two and three DL feature extractor models, based on accuracy across various classifiers, are combined into a deep feature ensemble, which is subsequently input into different ML classifiers for final predictions. Finally, we explored constructing an ensemble of the top two and top three ML classifiers by incorporating features derived from the top five individual pre-trained DL models.

\section{Proposed Methodology} 
\label{pm}

\subsection{Overview}

\begin{figure*}[!t]
     \centering
     \includegraphics[width=1.1\textwidth]{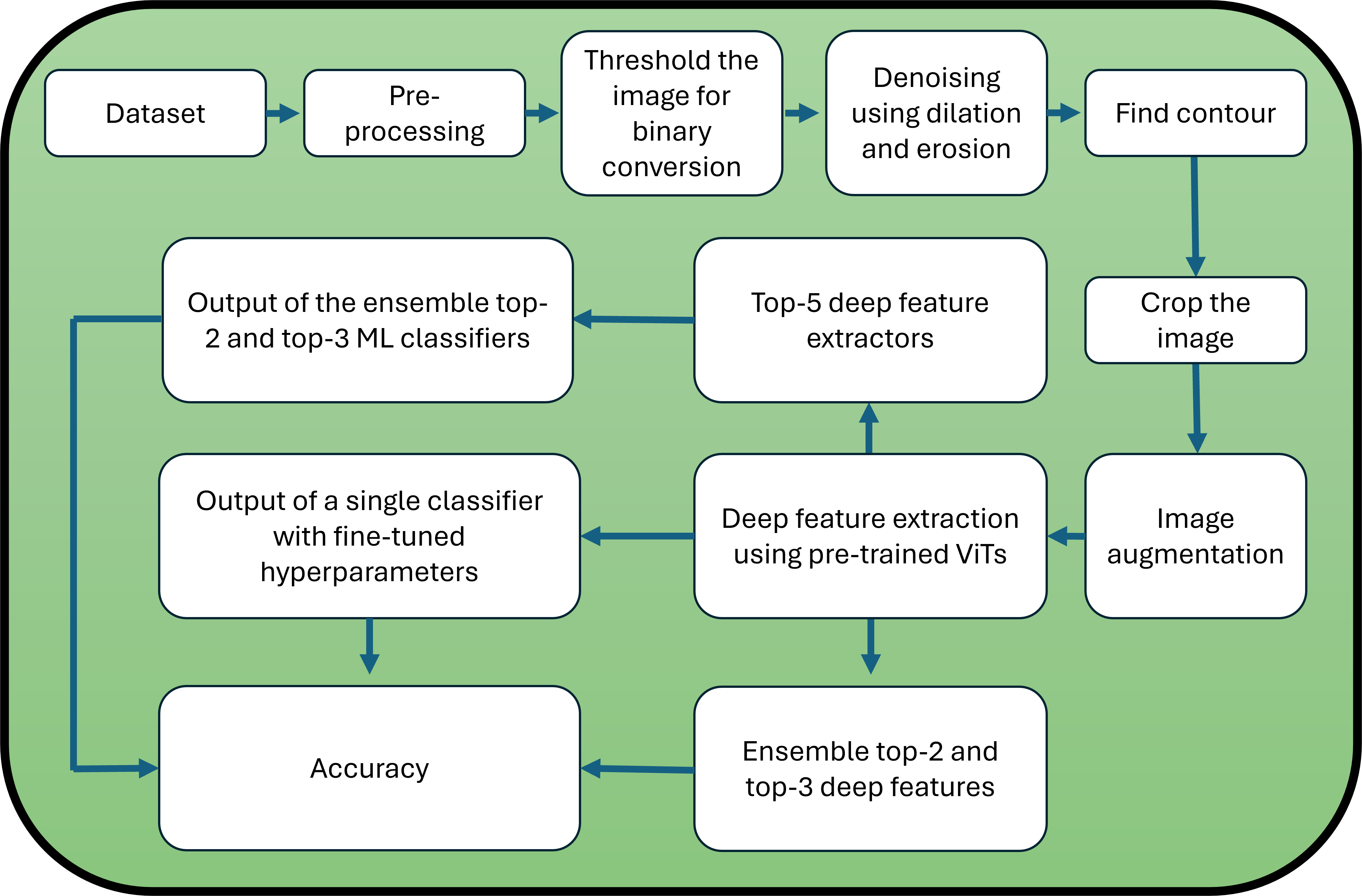}
     \caption{Workflow of our proposed double ensembling framework for brain tumor classification. }
     \label{workflow}
 \end{figure*}
\begin{figure*}[!ht]
     \centering
     \includegraphics[width=1\textwidth]{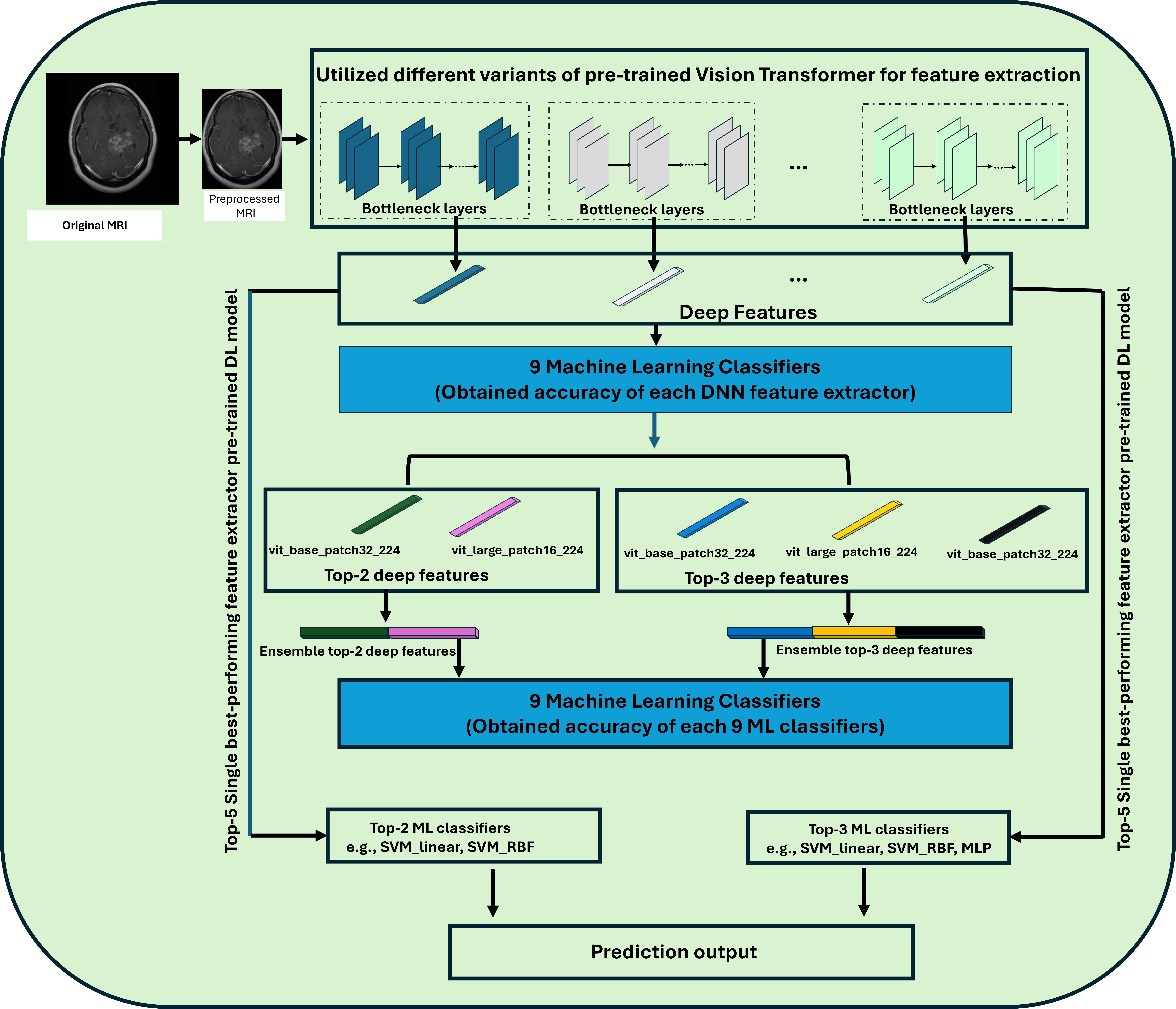}
     \caption{The proposed double ensembling architecture for brain tumor classification, designed to enhance predictive accuracy and robustness. This architecture incorporates a hierarchical ensemble of classifiers, leveraging diverse feature representations and combining their outputs to improve the model's ability to differentiate between normal and abnormal brain MRIs. }
     \label{proposed}
 \end{figure*}

As illustrated in Fig. \ref{proposed}, the proposed architecture for brain tumor classification utilizes a multi-faceted approach, integrating DL techniques with ML classifiers to achieve enhanced accuracy and robustness. As illustrated in Fig. \ref{workflow}, the process initiates with the meticulous preprocessing of MRIs, a crucial step that encompasses cropping to isolate the region of interest, resizing to standardize the input dimensions for subsequent processing, and augmentation to artificially expand the dataset and improve the model's generalization capability. These preprocessed images then serve as the input for pre-trained ViTs models, which are leveraged for their powerful feature extraction capabilities, capitalizing on transfer learning to adapt knowledge gained from large-scale image datasets to the specific task of brain tumor classification. The extracted features, representing high-level abstractions of the input images, are then subjected to a rigorous evaluation process using a diverse array of ML classifiers, allowing for a comparative analysis of their performance in distinguishing between normal and abnormal brain tumors.

Based on the empirical performance of the ML classifiers, a strategic selection process is undertaken to identify the top two or three deep feature sets, which are then combined within an ensemble module. This ensemble approach leverages the complementary strengths of different feature representations, potentially leading to a more comprehensive and discriminative feature space for classification. The concatenated features, representing a fusion of the most salient information extracted by the ViTs models, are then fed as input to another set of ML classifiers, which are tasked with making the final prediction regarding the type of brain tumor present in the input image. This cascaded approach, involving both deep feature extraction and ensemble learning, aims to maximize the accuracy and reliability of brain tumor classification, offering a potentially valuable tool for assisting medical professionals in diagnosis and treatment planning.

\subsection{Datasets}
\begin{figure*}[!ht]
     \centering
     \includegraphics[width=0.9\textwidth]{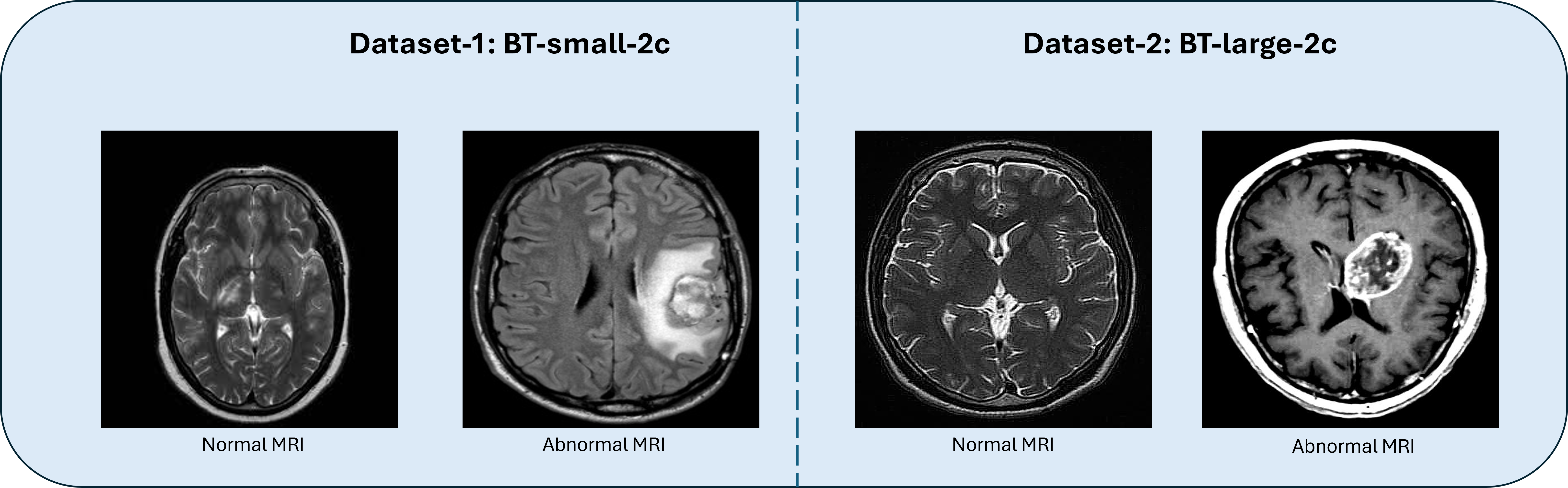}
     \caption{MRIs with Corresponding Classes from Each Dataset. }
     \label{datasetimg}
 \end{figure*}

In this study, we performed a series of experiments using two publicly available brain MRI datasets to classify brain tumors. The first dataset, termed BT-small-2c, was obtained from Kaggle \cite{chakrabarty2019brain} and includes 253 images, with 155 showing tumors and 98 without. The second dataset, labeled BT-large-2c, was also sourced from Kaggle and contains 3,000 images evenly distributed between 1,500 tumor-present and 1,500 normal cases \cite{Hamada2020}. These datasets provide a comprehensive basis for assessing brain tumor classification methods. Figure \ref{datasetimg} presents representative samples from each dataset, illustrating the diversity in imaging modalities and tumor characteristics. 

\begin{table*}[!t]
\centering
\caption{illustrates each dataset details.}
\begin{tabular}{cccc}
\hline
\textbf{Types} & \textbf{Number of classes} & \textbf{Training set} & \textbf{Test set} \\ \hline
BT-small-2c    & 2                          & 202                   & 51                \\
BT-large-2c    & 2                          & 2400                  & 600                          \\ \hline
\end{tabular}
\label{dataset}
\end{table*}

Both datasets were divided into a training set, representing 80\% of the data, and a test set, accounting for the remaining 20\%. Table \ref{dataset} summarizes the datasets used in our experiments, while Fig. \ref{datasetimg} showcases sample brain MRIs from the BT-small-2c and BT-large-2c datasets. These datasets collectively contribute to a comprehensive and diverse resource for assessing the performance of brain tumor classification methodologies, allowing for more robust and generalizable models to be developed.

\subsection{Pre-Processing}
\label{pp}
In our brain MRI datasets, many images contain extraneous areas that negatively impact classification accuracy. To enhance performance, it is crucial to crop these images, removing irrelevant regions while retaining only the essential data. We utilize the cropping technique outlined in \cite{zhang2020finding}, which relies on calculating extreme points.

\begin{figure*}[!ht]
     \centering
     \includegraphics[width=1\textwidth]{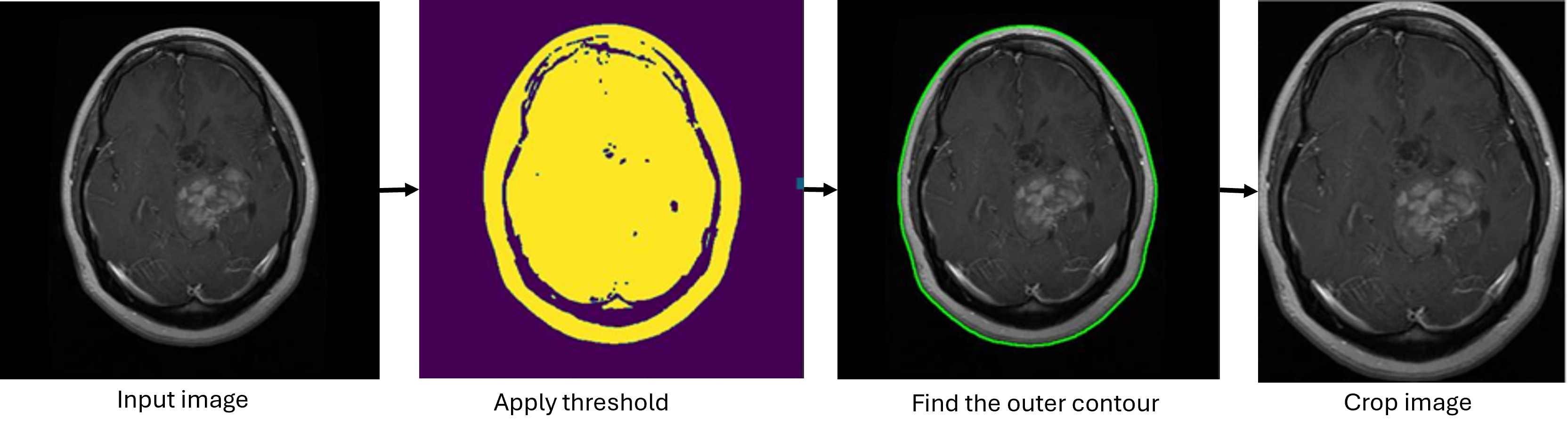}
     \caption{Pre-processing steps to crop the MRI images  }
     \label{preprocess}
 \end{figure*}

The cropping process for MRI images, illustrated in Figure \ref{preprocess}, begins by applying thresholding to convert the images into binary format, followed by dilation and erosion operations to reduce noise. Subsequently, the largest contour is identified in the binary images, and the four extreme points—topmost, bottommost, leftmost, and rightmost—are located. Using these points as a reference, the images are cropped and then resized using bicubic interpolation, which produces smoother curves and better manages the prominent edge noise often present in MRI images.

Furthermore, to address the limited size of our MRI dataset, we employed image augmentation, a technique that artificially increases a dataset by altering the original images. This approach creates multiple variations of each image by adjusting parameters such as scale, rotation, position, brightness, and other attributes. Studies have shown \cite{perez2017effectiveness,yang2022image} that augmenting datasets can enhance model classification accuracy more effectively than collecting additional data.

For our image augmentation process, we applied two specific methods: rotation and horizontal flipping. The rotation involved randomly rotating the input images by 90 degrees one or more times. Afterward, horizontal flipping was performed on each rotated image, further enriching the dataset with additional training samples.

\subsection{Deep Feature Extraction using Pre-trained Visions Transformers}
\label{dpe}
We employed variants of ViT-based models as DL feature extractors, utilizing their capacity to autonomously recognize and capture essential features without manual intervention. To address the limitations posed by the small size of our MRI dataset, we adopted a transfer learning approach for developing our feature extractor. Training and fine-tuning a DNN from scratch would be infeasible in this scenario. Instead, we leveraged the pre-trained fixed weights of each ViT model, originally trained on the large-scale ImageNet dataset, to efficiently extract deep features from brain MRI images. This strategy enhances the efficiency and reliability of our feature extraction process.

\begin{figure*}[!t]
     \centering
     \includegraphics[width=1\textwidth]{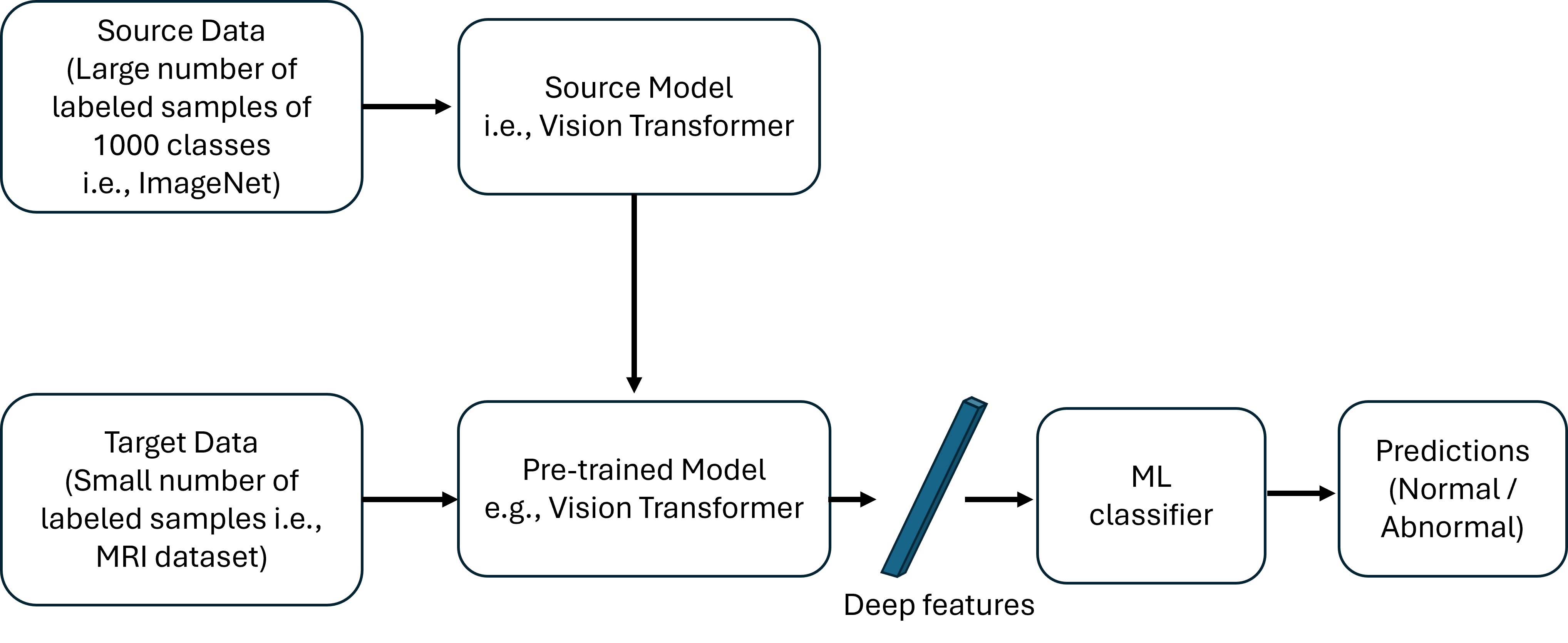}
     \caption{Methodology of transfer learning. }
     \label{tlearning}
 \end{figure*}

\begin{figure*}[h]
     \includegraphics[width=1\textwidth]{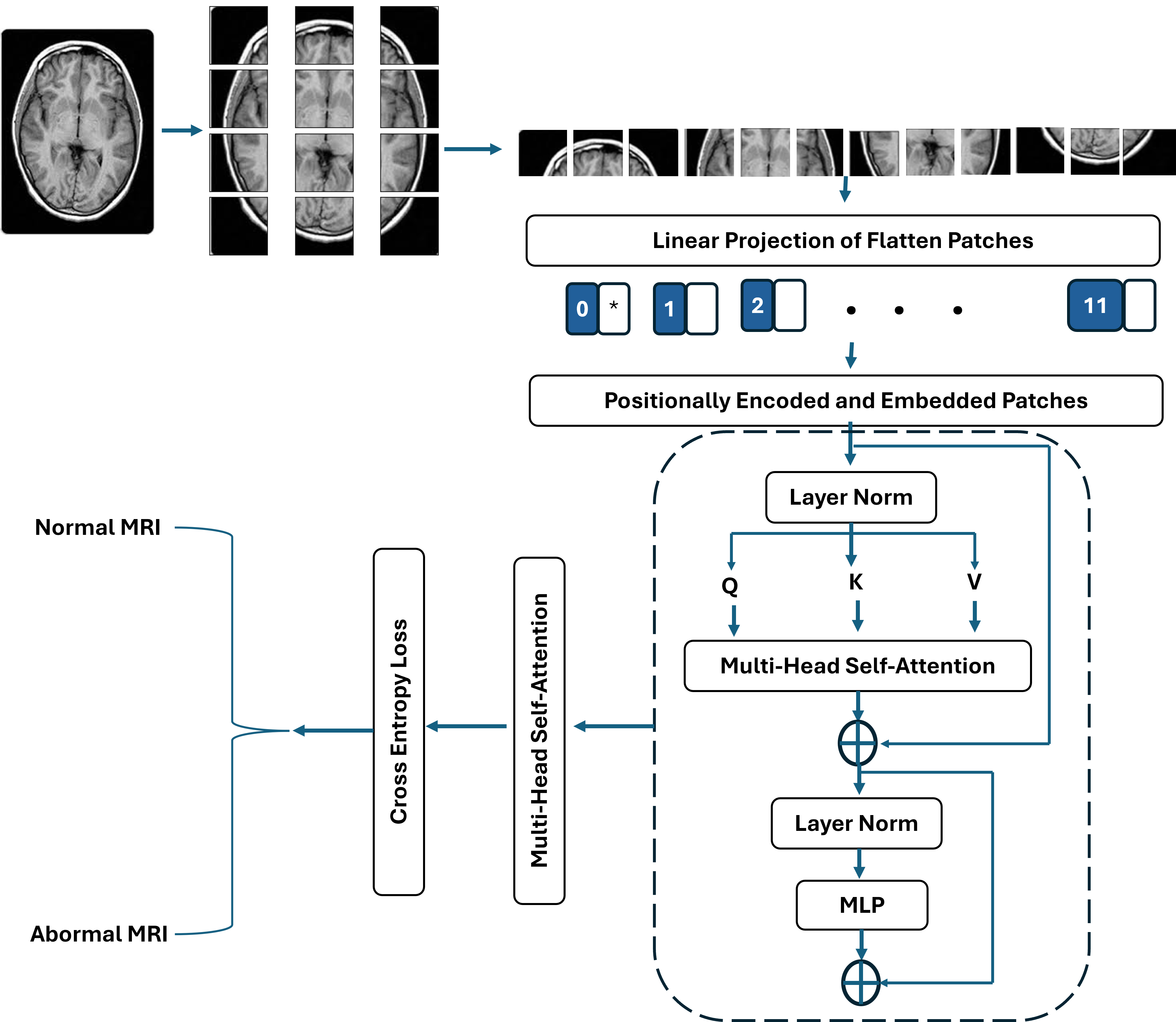}
     \caption{Overview of ViT Architecture for Brain MRI analysis. }
     \label{fig_vit_architecture}
 \end{figure*}

In this study, we utilized the following variants of ViT (vit\_base\_patch16\_224 , vit\_base\_patch32\_224, vit\_large\_patch16\_224, vit\_small\_patch32\_224, deit3\_small\_patch16\_224, vit\_base\_patch8\_224, vit\_tiny\_patch16\_224, vit\_small\_patch16\_224, vit\_base\_patch16\_384, vit\_tiny\_patch16\_384, vit\_small\_patch32\_384, vit\_small\_patch16\_384, vit\_base\_patch32\_384 \cite{dosovitskiy2020image, wu2020visual, deng2009imagenet}) models to extract the deep features from MRI as shown in Fig.  \ref{fig_vit_architecture}.

\subsection{Brain Tumor Classification using ML Classifiers} \label{MLcl}
Over the past few years, many studies have focused on traditional or classical ML techniques for brain tumor diagnosis. ML algorithms can aid physicians in interpreting medical imaging findings and reduce interpretation times. Over the past few years, many studies have focused on traditional or classical ML techniques for brain tumor diagnosis. ML algorithms can aid physicians in interpreting medical imaging findings and reduce interpretation times \cite{erickson2017machine}. 

To address the complexities inherent in brain tumor classification, this paper embarks on a comparative analysis of several ML classifiers, leveraging deep features extracted via pre-trained DL models. The integration of DL for feature extraction with traditional ML classifiers represents a powerful paradigm for medical image analysis, capitalizing on the strengths of both approaches. This strategy allows for the automated extraction of relevant image features, which are then fed into ML classifiers for categorization. By utilizing deep features extracted from pre-trained models, the system can leverage the knowledge learned from vast datasets, even when the available brain tumor dataset is limited in size. This approach is particularly advantageous in medical imaging, where acquiring large, annotated datasets can be challenging. The performance of these classifiers is rigorously evaluated and compared to determine their suitability for brain tumor classification, providing insights into the most effective approaches for this critical task.  These classifiers include neural networks with an MLP architecture, XGBoost, Gaussian Naïve Bayes, Adaptive Boosting, k-Nearest Neighbors (k-NN), RF, and SVMs with linear, sigmoid, and RBF kernels. The selection of these classifiers is motivated by their widespread use and proven effectiveness in various classification tasks, as well as their suitability for handling the types of data encountered in medical image analysis.  The implementation of these classifiers is facilitated by the scikit-learn ML library, a widely used and well-documented tool that provides a comprehensive suite of ML algorithms and utilities. The details of the ML classifiers and their hyper-parameter configurations applied in our brain tumor classification experiments are detailed in the subsequent sections, providing a comprehensive overview of the experimental setup and methodology.

\subsubsection{Multi layer Perceptron}

A Multilayer Perceptron (MLP) is a type of artificial neural network consisting of multiple layers of interconnected nodes (neurons), where each neuron is linked to those in the subsequent layers via directed edges. The main objective of an MLP is to establish a mapping between input data and corresponding outputs by iteratively refining its parameters using backpropagation \cite{meng2021agwo}. \textbf{Forward Propagation}: During forward propagation, outputs for each layer are computed based on the inputs, as well as the weights and biases associated with the neurons. Mathematically, the output $h^{(l)}$ of the $l$-th layer is computed as:

\begin{equation}
    y^{(l)} = w^{(l)}h^{l-1)} + b^{(l)},
\end{equation}

\begin{equation}
    h^{(l)} = ReLU(z^{(l)}),
\end{equation}

where $w^{(l)}$ and $b^{(l)}$ represents the weights and biases of the $l-$th layer, $y^{(l)}$ denotes the linear transformation of the input, and $h^{(l-1)}$ is the output from the previous layer and ReLU is utilized as an activation function defined as follows:
\begin{equation}
    ReLU(x) = max(0, x)
\end{equation}
The optimization of the MLP is carried out using the Adam optimizer, an adaptive learning rate method that combines the benefits of momentum and RMSProp. The weights and biases are updated iteratively using the following rules:

\begin{equation}
M_t=\beta_1 M_{t-1}+\left(1-\beta_1\right) \nabla L_t, \\
\end{equation}
\begin{equation}
    V_t=\beta_2 V_{t-1}+\left(1-\beta_2\right)\left(\nabla L_t\right)^2, \\
\end{equation}

\begin{equation}
\hat{M}_t=\frac{M_t}{1-\beta_1^t}, \quad \hat{V}_t=\frac{V_t}{1-\beta_2^t}, \\
\end{equation}

\begin{equation}
  \theta_{t+1}=\theta_t-\eta \frac{\hat{M}_t}{\sqrt{\hat{V}_t}+\epsilon},  
\end{equation}

where $M_t$ and $V_t$ represent the first and second moment estimates, respectively, $\nabla {L_t}$  denotes the gradient at time $t$, $\eta$ is the learning rate (set to 0.001) and $\beta_1$, $\beta_2,$ and $\epsilon$ are  hyperparameters of the optimizer. 

\textbf{Learning Process}: The learning process of the MLP involves minimizing a loss function $\textbf{L}$, which quantifies the difference between predicted outputs $\hat{y}$ and actual outputs $y$. For a dataset with $N$ samples, the loss is typically computed as the mean squared error (MSE) for regression tasks:

\begin{equation}
    L=\frac{1}{N} \sum_{i=1}^N\left(y_i-\hat{y}_i\right)^2
\end{equation}

The MLP progressively enhances its generalization and prediction accuracy by iteratively updating the weights and biases using the Adam optimizer. The integration of the ReLU activation function with the Adam solver facilitates efficient training and helps address challenges like vanishing gradients.

\subsubsection{Gaussian Naive Bayes}
This classifier is a ML model that assumes the features are conditionally independent given the class label. In this study, we utilize the Gaussian Naïve Bayes (NB) classifier as one of our approaches for brain tumor classification. With this classifier, the conditional probability \emph{P(y|X)} is calculated by multiplying the individual conditional probabilities, relying on the naïve assumption of independence among the features.

\begin{equation}
P(y|X) = \frac{P(y)P(X|y)}{P(X)} = \frac{P(y)
\prod_{i=1}^{n}P(x_i|y)}{P(X)}
\end{equation}

Here, {\emph{X}} represents a data instance derived from the deep features of a brain MR image, expressed as a feature vector \( (x_1, ..., x_n) \),. The variable {\emph{y}} denotes the target class—that is, the type of brain tumor—with two classes for the BT-small-2c and BT-large-2c MRI datasets. Since \( P(X) \) is the same for all classes, classification of a given instance is based on the remaining terms as follows:


\begin{equation}
\hat{Y} = \arg\max_Y \, P(Y) \left( \prod_{i=1}^{n} P(x_i \mid Y) \right)
\end{equation}

where \( (x_i|Y) \) is computed under the assumption that the feature likelihood follows a Gaussian distribution, as expressed below:

\begin{equation}
p(x_i|Y) = \frac{1}{\sqrt{2\pi \sigma_Y^2}}exp( \frac{(x_i-\mu_Y)^2}{2\sigma_Y^2} )
\end{equation}
where the parameters \( \mu_Y \) and \( \sigma_Y \) are determined via maximum likelihood estimation.

Here, the smoothing parameter, which denotes the fraction of the highest variance among all features added to the variances to ensure computational stability, is configured at \( 10^{-9}\), aligning with the default setting in the scikit-learn ML library.

\subsubsection{AdaBoost}

AdaBoost, introduced by \cite{freund1997decision}, is an ensemble learning algorithm designed to enhance overall performance by combining multiple classifiers. It constructs a strong classifier through an iterative process that assigns weights to individual weak classifiers, updating them during each boosting iteration. This approach trains on weighted data samples to enable the combined model to accurately predict class labels, such as distinguishing between binary datasets (BT-small-2c and BT-large-2c). Since any ML classifier supporting sample weights can serve as a base model, we chose the decision tree classifier due to its widespread use with AdaBoost. Additionally, we configured the algorithm to use 100 estimators.

\subsubsection{K-Nearest Neighbors}
The k-Nearest Neighbors (k-NN) algorithm is a simple yet effective classification technique that makes predictions directly based on the stored training data. For instance, when classifying a new data point (such as a deep feature extracted from a brain MRI), k-NN identifies the k closest training samples by calculating their distances from the new point. The algorithm then assigns the new data point the label most frequently occurring among its k nearest neighbors. This approach is applicable to both binary and multi-class classification tasks. Although Manhattan and Euclidean distances are commonly used for similarity measurement, our method utilizes the Euclidean distance metric. The Euclidean distance, \textit{d}, between two points \textit{x} and \textit{y} is calculated as follows:

\begin{equation}
d(X,Y) = \sqrt{ ( \sum\nolimits_{i=1}^{N}(X_i-Y_i)^2) }  
\end{equation}


\begin{equation}
    P = (Y_i),i,...,n
\end{equation}
where $Y_i$ represents a specific training sample in the training dataset, $n$ denotes the over all training samples, and $c_i$ corresponds to the class label associated with $Y_i$. 

\begin{itemize}
    \item During the testing phase, the distances between the new feature vector and the stored feature vectors from the training data are calculated. The new example is then classified based on the majority vote of its $k$-nearest neighbors.
\end{itemize}

The accuracy of the algorithm is evaluated during the testing phase using the correct classifications. If the performance is unsatisfactory, the value of $k$ can be adjusted to achieve a more acceptable level of accuracy. In this study, the neighbors size is varied from 1 to 4, and the value that resulted in the highest accuracy was selected.

\subsubsection{Random Forest}
RF introduced by Breiman \cite{breiman2001random}, is an ensemble learning algorithm that generates multiple decision trees using the bagging technique. It classifies new data points, such as deep features extracted from brain MRI images, into specific target categories. For the BT-small-2c and BT-large-2c datasets, RF differentiates between two classes.

During the construction of each tree, RF randomly selects $n$ features to identify the optimal split point, using the Gini index as the cost function. This random feature selection reduces correlations between trees, thereby improving the overall accuracy of the ensemble. To make predictions, the algorithm passes the input data through all the decision trees, with each tree producing a class prediction. The final class label is determined through majority voting, where the class with the highest number of votes is selected as the predicted label.

In this study, the square root of the total number of features was used to determine the optimal split criteria. Tree sizes ranging from 1 to 150 were evaluated, and the configuration that achieved the highest accuracy was selected.

\subsubsection{Support Vector Machine}
SVM \cite{cortes1995support}, introduced by Vladimir N. Vapnik, are robust classification algorithms that utilize kernel functions to transform input data into higher-dimensional feature spaces, facilitating the determination of optimal separating hyperplanes.

\begin{equation}
f(x_i) =  \sum_{n=1}^{N}\alpha_n y_n K(x_n, x_i)+b 
\end{equation}

Here, the support vectors, denoted as \(x_n\), represent the deep features extracted from brain MR images. The Lagrange multipliers, \(\alpha_n\), are coefficients assigned to each support vector during the optimization process. The target classes, \(y_n\), correspond to the classification labels in the datasets used in this study: two binary-class datasets (normal and abnormal) and one dataset with four classes, where \(n=1,2,3,..., N\).

We employed the most widely utilized kernel functions within the SVM algorithm: (1) the linear kernel, (2) the sigmoid kernel, and (3) the RBF kernel as outlined in Table \ref{svms}. Additionally, the SVM relies on two critical hyperparameters: C and Gamma. The C hyperparameter governs the soft margin cost function, determining the impact of each support vector, while Gamma influences the degree of curvature in the decision boundary. We tested Gamma values of `scale' and 'auto' with values [0.1, 1, and 10], and C values of [0.1, 1, 10, 100], ultimately choosing the Gamma and C combination that yielded the highest accuracy.

\begin{table}[!ht]
\centering
	\caption{Kernel types and their required parameters.}
	\label{tab:2}
	\setlength{\tabcolsep}{5.6mm}
	\begin{tabular}{cccc}
		\toprule
		\textbf{Kernel} & \textbf{Equation} & \textbf{Parameters } \\
		\midrule
		Linear & \(  K(x_n,x_i) = (x_n,x_i) \) & - \\
		Sigmoid & \( K(x_n,x_i) = tanh (\gamma (x_n,x_i)+C) \) & \( \gamma, C \)   \\
		RBF & \( K(x_n,x_i) = exp(-\gamma\left\|x_n-x_i\right\|^2+C) \) & \( \gamma, C \)  \\
		\bottomrule
	\end{tabular}
    \label{svms}
\end{table}

\subsection{Hyperparamter tuning on ML models}
HPO involves identifying the most effective set of hyperparameter values and the optimal arrangement for categorical hyperparameters. This process aims to improve model performance by reducing a predefined loss function, leading to more accurate results with fewer mistakes. Hyperparameter tuning \cite{yu2020hyper} refers to the method of crafting the perfect model structure with an ideal hyperparameter setup. Since each ML algorithm comes with its own unique hyperparameters, manually adjusting them demands a thorough knowledge of the models and their corresponding hyperparameter settings. Some proposed \cite{tran2020hyper} automated hyperparameter tuning techniques, such as random search, grid search, and Bayesian optimization, offer greater adaptability compared to conventional approaches for choosing the best hyperparameters, ultimately boosting model performance more efficiently.

ML tasks can be described as developing a model M that reduces a predefined loss function $L(X_{Ts}; M)$ on a specific test set $X_{Ts}$, where the loss function (L) represents the error rate. A learning algorithm A utilizes a training set $X_{Tr}$ to construct the model M, frequently addressing a nonconvex optimization challenge. This learning algorithm A incorporates certain hyperparameters lambda $\lambda$, and the model M is defined as $M = A(X_{Tr})$; $\lambda$. The goal of HPO is to identify the optimal settings $\lambda^{ast}$ that produce an ideal model $M^{ast}$, which minimizes the loss function $L(X_{Ts}; M)$.


\begin{align}
\lambda^* &= \underset{\lambda}{\operatorname{argmin}} \, L(X_{Ts}; A(X_{Tr}; \lambda)) &= \underset{\lambda}{\operatorname{argmin}} \, F(X_{Ts}, X_{Tr}, A, \lambda, L)
\end{align}

Here, F represents the model’s objective function, which takes $\lambda$—a set of hyperparameters—and returns the associated loss. The loss function $L$ and the learning algorithm are selected, and the datasets $X_{Ts}$ (test set) and $X_{Tr}$ (training set) are provided \cite{claesen2015hyperparameter}. These elements vary based on the chosen model, the hyperparameter search space, and the selected ML classifiers.

In this study, hyperparameter tuning was uniformly applied to nine ML models, with each model undergoing the same optimization process using grid search. This approach aimed to determine the optimal hyperparameters for each model, striking a balance between accuracy and computational efficiency. By maintaining consistency in the tuning process, the models were evaluated under comparable conditions, ensuring a fair assessment of their predictive performance and generalization capabilities.

As illustrated in Table \ref{MLparamtuning}, to enhance the effectiveness of our ML models, we perform HPO through grid search methods \cite{belete2022grid}. The hyperparameter settings that produce the best outcomes on the validation set are chosen for each model. Table \ref{without-pre-processing-BTlarge},  \ref{BT-large2c-fine-tune-all}, \ref{BT-small-finetune-all}, depict the results obtained from our ML models following this hyperparameter tuning process.

\begin{table*}[!ht]
\centering
\caption{Selected hyperparameter with search space for ML classifiers.}
\scalebox{0.6}{
\begin{tabular}{cccc}
\hline
\textbf{Model} & \textbf{Hyperparameter}                                                                                                                                                            & \textbf{Search Space}                                                                                                                                                                                                                                 & \textbf{Type}                                                                                                    \\ \hline
XGBoost        & \begin{tabular}[c]{@{}c@{}}max\_depth\\ learning\_rate\\ subsample\\ n\_estimators\end{tabular}                                                                                    & \begin{tabular}[c]{@{}c@{}}{[}3, 5, 7{]},\\ {[}0.1, 0.01, 0.001{]}\\ {[}0.5, 0.7, 1{]}\\ {[}100, 200, 300{]}\end{tabular}                                                                                                                             & \begin{tabular}[c]{@{}c@{}}Discrete\\ Continuous\\ Continuous\\ Discrete\end{tabular}                            \\ \hline
MLP            & \begin{tabular}[c]{@{}c@{}}hidden\_layer\_sizes\\ activation\\ solver\\ max\_iter\\ momentum\end{tabular}                                                                          & \begin{tabular}[c]{@{}c@{}}{[}(50,), (100,22), (100,100, 50), (100, 50, 36, 30), (100, 100, 200, 150, 100){]}\\ {[}relu, tanh, logistic{]}\\ {[}adam, sgd, lbfgs{]}\\ {[}1000{]}\\ {[}0.9, 0.95, 0.99{]}\end{tabular}                                 & \begin{tabular}[c]{@{}c@{}}Discrete\\ \\ Categorical\\ Discrete\\ Continuous\end{tabular}                        \\ \hline
Gaussian NB    & \begin{tabular}[c]{@{}c@{}}var\_smoothing\\ priors\end{tabular}                                                                                                                    & \begin{tabular}[c]{@{}c@{}}{[}1e-9, 1e-8, 1e-7, 1e-6, 1e-5{]}\\ {[}None, {[}0.3, 0.7{]}, {[}0.4, 0.6{]}, {[}0.5, 0.5{]}{]}\end{tabular}                                                                                                               & \begin{tabular}[c]{@{}c@{}}Continuous\\ Continuous\end{tabular}                                                  \\ \hline
Adaboost       & \begin{tabular}[c]{@{}c@{}}n\_estimators\\ learning\_rate\end{tabular}                                                                                                             & \begin{tabular}[c]{@{}c@{}}{[}50, 70, 90, 120, 180, 200{]}\\ {[}0.001, 0.01, 0.1, 1, 10{]}\end{tabular}                                                                                                                                               &                                                                                                                  \\ \hline
KNN            & \begin{tabular}[c]{@{}c@{}}n\_neighbors\\ weights\\ algorithm\\ leaf\_size\\ p\\ metric\\ n\_jobs\end{tabular}                                                                     & \begin{tabular}[c]{@{}c@{}}list(range(1, 31))\\ {[}uniform, distance{]}\\ {[}autom ball\_tree, kd\_tree, brute{]}\\ list(range(10, 51, 5))\\ {[}1, 2{]}\\ {[}euclidean, manhattan, minkowski{]}\\ {[}-1{]}\end{tabular}                               & \begin{tabular}[c]{@{}c@{}}Discrete\\ Categorical\\ \\ Discrete\\ Discrete\\ Categorical\\ Discrete\end{tabular} \\ \hline
RF             & \begin{tabular}[c]{@{}c@{}}n\_estimators\\ max\_depth\\ min\_samples\_split\\ min\_samples\_leaf\\ max\_features\\ bootstrap\\ criterion\\ oob\_score\\ random\_state\end{tabular} & \begin{tabular}[c]{@{}c@{}}{[}100, 200, 300, 400, 500{]}\\ {[}None, 10, 20, 30, 40, 50{]}\\ {[}2, 5, 10{]}\\ {[}1, 2, 4{]}\\ {[}auto, sqrt, log2{]}\\ {[}True, False{]}\\ {[}gini, entropy{]}\\ {[}True, False{]}\\ {[}42{]}\end{tabular}             & \begin{tabular}[c]{@{}c@{}}Discrete\\ Discrete\\ Discrete\\ Discrete\\ Categorical\end{tabular}                  \\ \hline
SVM\_linear    & \begin{tabular}[c]{@{}c@{}}C\\ kernel\\ tol\\ class\_weight\\ random\_state\end{tabular}                                                                                           & \begin{tabular}[c]{@{}c@{}}{[}0.1, 1, 10, 100, 1000{]}\\ {[}linear{]}\\ {[}1e-3, 1e-4, 1e-5{]}\\ {[}None, balanced{]}\\ {[}42{]}\end{tabular}                                                                                                         & \begin{tabular}[c]{@{}c@{}}Continuous\\ Categorical\\ \\ \\ Discrete\end{tabular}                                \\ \hline
SVM\_sigmoid   & \begin{tabular}[c]{@{}c@{}}kernel\\ C\\ gamma\\ coef0\\ tol\\ class\_weight\\ shrinking\\ probability\\ cache\_size\\ random\_state\end{tabular}                                   & \begin{tabular}[c]{@{}c@{}}sigmoid\\ {[}0.1, 1. 10, 100{]}\\ {[}scale, auto{]}\\ {[}0.0, 0.1, 0.5, 1.0{]}\\ {[}1e-3, 1e-4, 1e-5{]}\\ {[}None, balanced{]}\\ {[}True, False{]}\\ {[}True, False{]}\\ {[}200.0, 500.0, 100.0{]}\\ {[}42{]}\end{tabular} & \begin{tabular}[c]{@{}c@{}}Continuous\\ Categorical\\ Continuous\end{tabular}                                    \\ \hline
SVM\_RBF       & \begin{tabular}[c]{@{}c@{}}C\\ gamma\\ kernel\\ class\_weight\\ shrinking\\ probability\\ tol\\ cache\_size\\ max\_iter\end{tabular}                                               & \begin{tabular}[c]{@{}c@{}}{[}0.1, 1, 10, 100{]}\\ {[}scale, auto, 0.1, 1, 10{]}\\ {[}rbf{]}\\ {[}None, balanced{]}\\ {[}True, False{]}\\ {[}True, False{]}\\ {[}1e-3, 1e-4{]}\\ {[}200, 500, 1000{]}\\ {[}-1, 1000, 5000{]}\end{tabular}             & \begin{tabular}[c]{@{}c@{}}Discrete\\ Categorical\\ \\ \\ \\ \\ \\ Discrete\end{tabular}                         \\ \hline
\end{tabular}
}
\label{MLparamtuning}
\end{table*}

\begin{table}[!ht]
\centering
\caption{Accuracies of pre-trained ViT models with fine-tuned hyperparameters of ML classifiers on the non-preprocessed BT-large-2c dataset.}
\scalebox{0.55}{
\begin{tabular}{ccccccccccc}
\hline
\multirow{2}{*}{\textbf{\begin{tabular}[c]{@{}c@{}}Deep Feature from the\\ Pre-trained ViT models\end{tabular}}} & \multicolumn{10}{c}{\textbf{ML Classifier Accuracy}}                                                                                                                                        \\ \cline{2-11} 
                                                                                                                 & \textbf{XGBoost} & \textbf{MLP}    & \textbf{GaussianNB} & \textbf{Adaboost} & \textbf{KNN} & \textbf{RFClassifier} & \textbf{SVM\_linear} & \textbf{SVM\_sigmoid} & \textbf{SVM\_RBF} & \textbf{Average} \\ \hline
vit\_base\_patch16\_224                                                                                          & 0.9001           & 0.9800          & 0.8515              & 0.9731            & 0.9652       & 0.9752                & 0.9705               & 0.9200                & 0.9712            & 0.9452           \\
vit\_base\_patch32\_224                                                                                          & 0.8832           & 0.9771          & 0.8600              & 0.9785            & 0.9532       & 0.9652                & 0.9800               & 0.9441                & 0.9804            & 0.9469           \\
vit\_large\_patch16\_224                                                                                         & 0.9221           & 0.9851          & 0.8536              & 0.9855            & 0.9702       & 0.9632                & 0.9831               & 0.9591                & 0.9732            & \textbf{0.9550}  \\
vit\_small\_patch32\_224                                                                                         & 0.8911           & 0.9728          & 0.8752              & 0.9623            & 0.9723       & 0.9501                & 0.9506               & 0.9199                & 0.9800            & 0.9416           \\
deit3\_small\_patch16\_224                                                                                       & 0.8544           & 0.9900          & 0.7857              & 0.9513            & 0.9602       & 0.9451                & 0.9401               & 0.8408                & 0.9766            & 0.9160           \\
vit\_base\_patch8\_224                                                                                           & 0.9121           & 0.9917          & 0.8469              & 0.9641            & 0.9532       & 0.9502                & 0.9700               & 0.8600                & 0.9802            & 0.9365           \\
vit\_tiny\_patch16\_224                                                                                          & 0.9016           & 0.9850          & 0.8321              & 0.9607            & 0.9700       & 0.9500                & 0.9415               & 0.8722                & 0.9612            & 0.9305           \\
vit\_small\_patch16\_224                                                                                         & 0.9024           & 0.9900          & 0.8400              & 0.9802            & 0.9739       & 0.9700                & 0.9602               & 0.9192                & 0.9709            & 0.9452           \\
vit\_base\_patch16\_384                                                                                          & 0.9132           & 0.9680          & 0.8768              & 0.9700            & 0.9621       & 0.9699                & 0.9631               & 0.9504                & 0.9700            & 0.9493           \\
vit\_tiny\_patch16\_384                                                                                          & 0.9056           & 0.9666          & 0.7854              & 0.9703            & 0.9700       & 0.9708                & 0.9523               & 0.8700                & 0.9904            & 0.9313           \\
vit\_small\_patch32\_384                                                                                         & 0.9214           & 0.9835          & 0.8899              & 0.9733            & 0.9833       & 0.9667                & 0.9612               & 0.9212                & 0.9808            & 0.9535           \\
vit\_small\_patch16\_384                                                                                         & 0.8932           & 0.9627          & 0.8244              & 0.9621            & 0.9801       & 0.9700                & 0.9510               & 0.9219                & 0.9803            & 0.9384           \\
vit\_base\_patch32\_384                                                                                          & 0.9021           & 0.9732          & 0.8241              & 0.9800            & 0.9800       & 0.9623                & 0.9701               & 0.9505                & 0.9612            & 0.9448           \\ \hline
Average                                                                                                          & 0.9002           & \textbf{0.9789} & 0.8420              & 0.9701            & 0.9687       & 0.9622                & 0.9611               & 0.9115                & 0.9751            &                  \\ \hline
\end{tabular}
}
\label{without-pre-processing-BTlarge}
\end{table}

\begin{table}[!ht]
\centering
\caption{Accuracies of pre-trained ViT models using fine tune hyperparameter of ML classifiers on preprocessed BT-large-2c dataset. The top-3 deep features were represented using \(\star \).}
\scalebox{0.55}{
\begin{tabular}{cllllllllll}
\hline
\multirow{2}{*}{\textbf{\begin{tabular}[c]{@{}c@{}}Deep Feature from the\\ Pre-Trained ViT Model\end{tabular}}} & \multicolumn{9}{c}{\textbf{ML Classifier Accuracy}}                                                                                                                      &                  \\ \cline{2-11} 
                                                                                                                & \textbf{XGBoost} & \textbf{MLP}    & \textbf{GaussianNB} & \textbf{Adaboost} & \textbf{KNN} & \textbf{RFClassifier} & \textbf{SVM\_linear} & \textbf{SVM\_sigmoid} & \textbf{SVM\_RBF} & \textbf{Average} \\ \cline{2-11} 
vit\_base\_patch16\_224                                                                                         & 0.9483           & 0.9917          & 0.865               & 0.9817            & 0.9867       & 0.98                  & 0.9917               & 0.9383                & 0.99              & 0.9637           \\
\textbf{vit\_base\_patch32\_224\(\star \)}                                                                                & 0.9517           & 0.995           & 0.8767              & 0.9883            & 0.9883       & 0.9783                & 0.9917               & 0.9633                & 0.995             & \textbf{0.9698}  \\
\textbf{vit\_large\_patch16\_224\(\star \)}                                                                               & 0.96             & 0.995           & 0.8683              & 0.9933            & 0.985        & 0.9833                & 0.9967               & 0.9833                & 0.9967            & \textbf{0.9735}  \\
vit\_small\_patch32\_224                                                                                        & 0.93             & 0.9867          & 0.8917              & 0.9833            & 0.9933       & 0.9683                & 0.9717               & 0.9367                & 0.995             & 0.9619           \\
deit3\_small\_patch16\_224                                                                                      & 0.8717           & 0.9867          & 0.7983              & 0.96              & 0.97         & 0.9517                & 0.955                & 0.855                 & 0.99              & 0.9265           \\
vit\_base\_patch8\_224                                                                                          & 0.94             & 0.995           & 0.855               & 0.9833            & 0.985        & 0.9733                & 0.995                & 0.8817                & 0.9933            & 0.9557           \\
vit\_tiny\_patch16\_224                                                                                         & 0.91             & 0.9867          & 0.865               & 0.9767            & 0.985        & 0.9717                & 0.9533               & 0.895                 & 0.9867            & 0.9478           \\
vit\_small\_patch16\_224                                                                                        & 0.94             & 0.9917          & 0.8583              & 0.9917            & 0.985        & 0.98                  & 0.975                & 0.9383                & 0.9933            & 0.9615           \\
vit\_base\_patch16\_384                                                                                         & 0.94             & 0.9883          & 0.895               & 0.9867            & 0.9867       & 0.9783                & 0.985                & 0.9633                & 0.985             & 0.9676           \\
vit\_tiny\_patch16\_384                                                                                         & 0.9167           & 0.9917          & 0.8083              & 0.98              & 0.9883       & 0.9833                & 0.9733               & 0.8983                & 0.9933            & 0.9481           \\
vit\_small\_patch32\_384                                                                                        & 0.9433           & 0.99            & 0.9067              & 0.9883            & 0.99         & 0.9867                & 0.975                & 0.9483                & 0.9967            & 0.9694           \\
vit\_small\_patch16\_384                                                                                        & 0.9367           & 0.9883          & 0.8333              & 0.9817            & 0.99         & 0.99                  & 0.975                & 0.945                 & 0.9917            & 0.9591           \\
\textbf{vit\_base\_patch32\_384\(\star \)}                                                                                & 0.9517           & 0.9933          & 0.8833              & 0.9917            & 0.985        & 0.9883                & 0.99                 & 0.965                 & 0.995             & \textbf{0.9715}  \\ \hline
Average                                                                                                         & 0.9338           & \textbf{0.9908} & 0.8619              & 0.9836            & 0.986        & 0.9779                & 0.9791               & 0.9317                & 0.9924            &                  \\ \hline
\end{tabular}
}
\label{BT-large2c-fine-tune-all}
\end{table}

\begin{table}[!hbt]
\centering
\caption{Accuracies of pre-trained ViT models using fine-tune hyperparameter of ML classifiers on BT-small-2c dataset. The top-3 deep features were represented using \(\star \).}
\scalebox{0.6}{
\begin{tabular}{cccccccccc}
\hline
\multirow{2}{*}{\textbf{\begin{tabular}[c]{@{}c@{}}Deep Feature from the\\ Pre-Trained ViT Model\end{tabular}}} & \multicolumn{9}{c}{\textbf{ML Classifier Accuracy}}                                                                                                                   \\ \cline{2-10} 
                                                                                                                & \textbf{MLP} & \textbf{GaussianNB} & \textbf{Adaboost} & \textbf{KNN} & \textbf{RFClassifier} & \textbf{SVM\_linear} & \textbf{SVM\_sigmoid} & \textbf{SVM\_RBF} & \textbf{Average} \\ \cline{2-10} 
vit\_base\_patch16\_224 \(\star \)                                                                                         & 0.9750       & 0.8944              & 0.9500            & 0.9105       & 0.9250                & 0.9750               & 0.9339                & 0.9750            & \textbf{0.9423}  \\
vit\_base\_patch32\_224                                                                                         & 0.9000       & 0.8460              & 0.9000            & 0.9355       & 0.9589                & 0.9016               & 0.9339                & 0.9339            & 0.9137           \\
vit\_large\_patch16\_224                                                                                        & 0.9000       & 0.8460              & 0.8089            & 0.8194       & 0.9000                & 0.9750               & 0.9750                & 0.9750            & 0.8999           \\
vit\_small\_patch32\_224 \(\star \)                                                                                       & 0.9500       & 0.9105              & 0.9339            & 0.9105       & 0.9339                & 0.9500               & 0.9589                & 0.9750            & \textbf{0.9403}  \\
deit3\_small\_patch16\_224                                                                                      & 0.8516       & 0.7032              & 0.8250            & 0.8371       & 0.8500                & 0.9427               & 0.7427                & 0.9016            & 0.8318           \\
vit\_base\_patch8\_224                                                                                          & 0.9500       & 0.8782              & 0.9089            & 0.8855       & 0.9000                & 0.9339               & 0.9339                & 0.9500            & 0.9175           \\
vit\_tiny\_patch16\_224                                                                                         & 0.9589       & 0.8944              & 0.8839            & 0.9032       & 0.9250                & 0.8516               & 0.9427                & 0.9750            & 0.9168           \\
vit\_small\_patch16\_224 \(\star \)                                                                                        & 0.9750       & 0.8282              & 0.9750            & 0.9016       & 0.9500                & 0.9339               & 0.9750                & 0.9750            & \textbf{0.9392}           \\
vit\_base\_patch16\_384                                                                                         & 0.9339       & 0.9032              & 0.9500            & 0.8855       & 0.9500                & 0.9750               & 0.9339                & 0.9589            & 0.9363           \\
vit\_tiny\_patch16\_384                                                                                         & 0.9266       & 0.8032              & 0.8839            & 0.8121       & 0.9250                & 0.8194               & 0.9339                & 0.9589            & 0.8829           \\
vit\_small\_patch32\_384                                                                                        & 0.9427       & 0.9089              & 0.9339            & 0.9427       & 0.9339                & 0.9355               & 0.9339                & 0.9266            & 0.9323           \\
vit\_small\_patch16\_384                                                                                        & 0.9500       & 0.8766              & 0.9500            & 0.8121       & 0.9177                & 0.9177               & 0.9589                & 0.9750            & 0.9198           \\
vit\_base\_patch32\_384                                                                                         & 0.8750       & 0.8137              & 0.9089            & 0.9016       & 0.9250                & 0.9750               & 0.9339                & 0.9750            & 0.9135           \\ \hline
Average                                                                                                         & 0.9299       & 0.8543              & 0.9086            & 0.8813       & 0.9226                & 0.9297               & 0.9300                & \textbf{0.9581}   &                  \\ \hline
\end{tabular}
}
\label{BT-small-finetune-all}
\end{table}

\section{Experimental setup}
\label{experimental}
The experiments in this study were designed to evaluate the performance of the proposed hybrid approach for brain tumor classification. Details of the experimental setup are provided in the respective subsections.

\subsection{Implementation details}

In this study, we utilized 13 pre-trained ViT-based models as feature extractors, as described in Section \ref{dpe}. These models, pre-trained on the ImageNet dataset \cite{krizhevsky2012imagenet}, were adapted for our task by freezing the weights of their bottleneck layers. This approach preserves the learned features while preventing overfitting to our comparatively smaller dataset. Additionally, we incorporated nine distinct ML classifiers, detailed in Section \ref{MLcl}, to evaluate the extracted features. The combined use of diverse feature extraction techniques and classifiers ensures robust and comprehensive analysis of the MRI datasets. Before training, all input images underwent preprocessing steps as outlined in Section \ref{pp}. These steps included image cropping, resizing, and augmentation to enhance the dataset's quality and diversity, ultimately improving model performance.

All experiments were conducted using an NVIDIA RTX 3090 GPU, ensuring efficient processing and the ability to handle the computational demands of training DL models and performing extensive experimentation.

\subsubsection{Deep Feature Evaluation and Selection}
In this study, we evaluated the deep features extracted from 13 pre-trained ViT models using various ML classifiers, as outlined in Section \ref{MLcl}. The goal was to identify the top-performing deep features for each of the two MRI datasets based on their classification accuracy.

\subsubsection{Evaluation Criteria}
The evaluation process involved calculating the average accuracy achieved by each ViT model across nine different ML classifiers. This approach ensured a comprehensive assessment of the models' performance across a diverse set of classifiers, providing a robust basis for selection. To resolve ties where two or more models achieved identical accuracy, we prioritized models with lower standard deviation. A lower standard deviation indicates more consistent performance across different subsets of the data, which is critical for reliable classification.

\subsubsection{Rationale for Selection}
The primary reason for selecting only the top three deep features is to minimize redundancy and enhance diversity within the feature ensemble. Deep features extracted from closely related models often occupy overlapping feature spaces, which can lead to diminished ensemble performance due to lack of variability. By focusing on the top three models with distinct and reliable performance, we ensure the ensemble leverages complementary feature representations. The top three deep features are then utilized in our ensemble module, which is detailed in the subsequent subsection.

\subsubsection{Ensemble of Deep Features}
Ensemble learning is a powerful technique aimed at improving model performance and reducing the risks associated with relying on a single feature or model. By combining multiple features from various models into a unified predictive feature set, ensemble learning enhances the robustness and accuracy of classification tasks. This technique can be broadly classified into two categories: feature ensemble and classifier ensemble, depending on the level at which integration occurs.

\subsection{Feature Ensemble vs. Classifier Ensemble}
\begin{itemize}
    \item \textbf{Feature Ensemble}: In this approach, feature sets extracted from different models are combined into a single unified feature sequence. This integrated feature set is then fed into a classifier to produce the final prediction. The feature ensemble approach leverages the rich, diverse information embedded in the feature sets, offering a more detailed representation of the input data (e.g., MR images) compared to individual classifier outputs.

\item  \textbf{Classifier Ensemble}: This method combines the predictions from multiple classifiers, using techniques such as majority voting or weighted voting to determine the final output. 
\end{itemize}


\subsection{Feature ensemble methodology}
In the feature ensemble approach, we integrate deep features extracted from the top-2 and top-3 distinct pre-trained ViT models. This integration is achieved by concatenating the feature vectors into a single unified feature sequence. For instance, if vit\_base\_patch\_16\_224, vit\_small\_patch\_32\_224, and vit\_small\_patch\_16\_224 are identified as the top-performing models, their respective feature sets are concatenated to form a comprehensive feature vector. This combined feature set is then input into ML classifiers to predict the final output.

To assess the impact of combining multiple feature sets, we conducted a comparative analysis:
\begin{itemize}
    \item \textbf{Two-Feature Combinations:} We explored all possible pairwise combinations of deep features extracted from the top three models. Each pair was fed into the classifiers to evaluate performance.
    \item \textbf{Three-Feature Ensemble:} We integrated deep features from all three top-performing models and input the resulting concatenated feature vector into the classifiers.
\end{itemize}

This comparative analysis allowed us to evaluate the relative effectiveness of the three-feature ensemble against pairwise combinations, providing insights into the optimal configuration for robust classification.

\subsubsection{Classifier Ensemble Methodology}
In addition to feature-level integration, we also explored classifier ensembles by combining the predictions from multiple classifiers. Using techniques such as majority voting, the outputs of individual classifiers were aggregated to determine the final prediction. This approach provides an additional layer of robustness by leveraging the strengths of multiple classifiers.

\subsubsection{Insights and Advantages}
\begin{itemize}
    \item \textbf{Improved Representation:} The feature ensemble approach ensures that the classifier receives a comprehensive and detailed input, enhancing its ability to make accurate predictions.

\item \textbf{Increased Diversity:} By combining features from distinct models, the ensemble minimizes redundancy and captures diverse aspects of the input data.

\item \textbf{Robustness:} The combined use of feature and classifier ensembles enhances the overall reliability of the system, addressing potential weaknesses in individual models or classifiers.
\end{itemize}

The ensemble strategies employed in this study are integral to achieving superior performance in brain tumor classification tasks, as detailed in the experimental results.

\section{Experimental Results}\label{results}
The experimental results were derived from two publicly available datasets for brain tumor classification tasks, referred to as BT-small-2c and BT-large-2c. The experiments were conducted to evaluate the effectiveness of the proposed hybrid approach using a systematic methodology. The study was divided into three distinct experiments, each targeting a specific aspect of the classification task.

\subsection{Experiment1: Hyperparameter Tuning of ML Classifiers}
The first experiment focused on optimizing the performance of the nine ML classifiers employed in this study as illustrated in Table \ref{MLparamtuning}. Hyperparameter tuning was conducted for each classifier to enhance its ability to classify brain tumors accurately. This process involved adjusting key parameters for each ML model, such as the number of neighbors in k-NN, the number of estimators in ensemble methods (e.g., RF and AdaBoost), and the learning rate in gradient-boosting classifiers. The tuned hyperparameters significantly improved the classifiers' performance. For instance:

\begin{itemize}
    \item \textbf{k-NN:} The optimal value of k was determined to achieve the best balance between accuracy and computational efficiency.

\item \textbf{Random Forest:} The number of estimators and the maximum depth of trees were adjusted to maximize classification accuracy.

\item \textbf{AdaBoost:} The learning rate and the number of weak classifiers were fine-tuned for optimal results.
\end{itemize}

The accuracy was recorded for each classifier after tuning. These results formed the baseline for subsequent experiments.

\subsection{Experiment 2: Ensembling Deep Features with ML Classifiers}
The second experiment aimed to demonstrate the advantages of combining deep features extracted from the top two or three pre-trained ViT models with various ML classifiers. The top-performing deep features were identified in the feature evaluation stage based on their average accuracy across the tuned ML classifiers.

\subsubsection{Feature ensemble creation:}
\begin{itemize}
    \item \textbf{Top-2 Models:} Deep features from the top two ViT models were concatenated into a single feature vector and fed into each ML classifier.

\item \textbf{Top-3 Models:} Similarly, deep features from the top three models were combined and input into the classifiers.
\end{itemize}

\subsubsection{Performance comaparison:}
\begin{itemize}
    \item The classifiers were evaluated using the combined feature sets, and their performance was compared to that achieved using individual deep features.

\item Results showed that the feature ensembles provided a richer representation of the input data, leading to improved classification accuracy across most ML classifiers.
\end{itemize}

\subsection{Experiment 3: Ensembling ML Classifiers with Preprocessing Variants}
The final experiment explored the ensembling of ML classifiers to further enhance classification performance. This involved combining the predictions of the top two and top three ML classifiers using various preprocessing strategies:

\textbf{1. Enhance ensemble:}
\begin{itemize}
    \item Predictions from the top classifiers were combined using a simple majority voting mechanism. The optimal DL feature extractors and ML classifiers demonstrate dataset-specific performance characteristics. As illustrated in Table \ref{BT-small-top3-DL-model-simple-version}, on the BT-small-2c dataset, the ensemble of ``vit base patch16 224 + vit small patch32 224" in conjunction with an MLP classifier achieves a notable accuracy of 0.9589\%. In contrast, in Table \ref{BT-large-2c-top3-DL-simple-version}, the BT-large-2c dataset benefits most from the combination of ``vit large patch16 224 + vit base patch32 384" also paired with an MLP classifier, resulting in a superior accuracy of 0.9983\%. These results underscore the importance of tailoring the feature extraction and classification approach to the specific characteristics of the dataset to maximize performance."
\end{itemize}

\begin{table}[!ht]
\centering
\caption{Accuracies of top three pre-trained ViT models using a simple version on the BT-small-2c dataset. The table compares different deep features from the pre-trained ViT models in combination with various classification algorithms. }
\scalebox{0.5}{
\begin{tabular}{ccccccccc}
\hline
\multirow{2}{*}{\textbf{\begin{tabular}[c]{@{}c@{}}Deep Feature from the\\ Pre-Trained ViT Model\end{tabular}}}            & \multicolumn{8}{c}{\textbf{ML Classifier Accuracy}}                                                                                                                                                                                                                                                                \\ \cline{2-9} 
                                                                                                                           & \multicolumn{1}{l}{\textbf{MLP}} & \multicolumn{1}{l}{\textbf{GaussianNB}} & \multicolumn{1}{l}{\textbf{Adaboost}} & \multicolumn{1}{l}{\textbf{KNN}} & \multicolumn{1}{l}{\textbf{RFClassifier}} & \multicolumn{1}{l}{\textbf{SVM\_linear}} & \multicolumn{1}{l}{\textbf{SVM\_sigmoid}} & \multicolumn{1}{l}{\textbf{SVM\_RBF}} \\ \hline
vit\_base\_patch16\_224   + vit\_small\_patch32\_224                                                                       & 0.9589                           & 0.6129                                  & 0.9250                                & 0.9177                           & 0.9250                                    & 0.9750                                   & 0.9589                                    & 0.9750                                \\
vit\_base\_patch16\_224 +   vit\_small\_patch16\_224                                                              & 0.9750                           & 0.5968                                  & 0.8266                                & 0.8694                           & 0.8589                                    & 0.9750                                   & 0.9750                                    & 0.9750                                \\
vit\_small\_patch32\_224 +   vit\_small\_patch16\_224                                                             & 0.9750                           & 0.7427                                  & 0.9177                                & 0.8782                           & 0.9500                                    & 0.9750                                   & 0.9589                                    & 0.9750                                \\
\begin{tabular}[c]{@{}c@{}}vit\_base\_patch16\_224 +   vit\_small\_patch32\_224 + \\ vit\_small\_patch16\_224\end{tabular} & 0.9750                           & 0.5427                                  & 0.9589                                & 0.9266                           & 0.8782                                    & 0.9750                                   & 0.9750                                    & 0.9750                                \\ \hline
\end{tabular}
}
\label{BT-small-top3-DL-model-simple-version}
\end{table}

\begin{table}[!ht]
\centering
\caption{Accuracies of top three pre-trained ViT models using a simple version on the BT-large-2c dataset. }
\scalebox{0.5}{
\begin{tabular}{ccccccccc}
\hline
\multirow{2}{*}{\textbf{\begin{tabular}[c]{@{}c@{}}Deep Feature from the\\ Pre-Trained ViT Model\end{tabular}}}           & \multicolumn{8}{c}{\textbf{ML Classifier Accuracy}}                                                                                                                                                                                                                                                                \\ \cline{2-9} 
                                                                                                                          & \multicolumn{1}{l}{\textbf{MLP}} & \multicolumn{1}{l}{\textbf{GaussianNB}} & \multicolumn{1}{l}{\textbf{Adaboost}} & \multicolumn{1}{l}{\textbf{KNN}} & \multicolumn{1}{l}{\textbf{RFClassifier}} & \multicolumn{1}{l}{\textbf{SVM\_linear}} & \multicolumn{1}{l}{\textbf{SVM\_sigmoid}} & \multicolumn{1}{l}{\textbf{SVM\_RBF}} \\ \hline
vit\_large\_patch16\_224   + vit\_base\_patch32\_384                                                                      & 0.9983                           & 0.6700                                  & 0.9900                                & 0.9883                           & 0.9900                                    & 0.9967                                   & 0.9700                                    & 0.9967                                \\
vit\_large\_patch16\_224 +   vit\_base\_patch32\_224                                                                      & 0.9983                           & 0.6550                                  & 0.9883                                & 0.9917                           & 0.9717                                    & 0.9983                                   & 0.9800                                    & 0.9967                                \\
vit\_base\_patch32\_384 +   vit\_base\_patch32\_224                                                                       & 0.9950                           & 0.6983                                  & 0.9917                                & 0.9900                           & 0.9750                                    & 0.9950                                   & 0.9800                                    & 0.9967                                \\
\begin{tabular}[c]{@{}c@{}}vit\_large\_patch16\_224 +   vit\_base\_patch32\_384 +\\  vit\_base\_patch32\_224\end{tabular} & 0.9950                           & 0.6683                                  & 0.9883                                & 0.9900                           & 0.9833                                    & 0.9983                                   & 0.9867                                    & 0.9967                                \\ \hline
\end{tabular}
}
\label{BT-large-2c-top3-DL-simple-version}
\end{table}

\textbf{2. Ensemble with Enhance processing:}
The following preprocessing techniques were applied before ensembling:

\begin{table}[!ht]
\centering
\caption{Classification accuracies of the top three pre-trained ViT models using normalization and PCA for deep feature extraction, evaluated with various ML classifiers on the BT-small-2c dataset. }
\scalebox{0.5}{
\begin{tabular}{ccccccccc}
\hline
\multirow{2}{*}{\textbf{\begin{tabular}[c]{@{}c@{}}Deep Feature from the\\ Pre-Trained ViT Model\end{tabular}}}            & \multicolumn{8}{c}{\textbf{ML Classifier Accuracy}}                                                                                                                                                                                                                                                                \\ \cline{2-9} 
                                                                                                                           & \multicolumn{1}{l}{\textbf{MLP}} & \multicolumn{1}{l}{\textbf{GaussianNB}} & \multicolumn{1}{l}{\textbf{Adaboost}} & \multicolumn{1}{l}{\textbf{KNN}} & \multicolumn{1}{l}{\textbf{RFClassifier}} & \multicolumn{1}{l}{\textbf{SVM\_linear}} & \multicolumn{1}{l}{\textbf{SVM\_sigmoid}} & \multicolumn{1}{l}{\textbf{SVM\_RBF}} \\ \hline
vit\_base\_patch16\_224   + vit\_small\_patch32\_224                                                                       & 0.9339                           & 0.6129                                  & 0.9589                                & 0.9177                           & 0.8177                                    & 0.9750                                   & 0.9589                                    & 0.9750                                \\
vit\_base\_patch16\_224 +   vit\_small\_patch16\_224                                                                       & 0.9750                           & 0.5645                                  & 0.8516                                & 0.8694                           & 0.7750                                    & 0.9750                                   & 0.9750                                    & 0.9750                                \\
vit\_small\_patch32\_224 +   vit\_small\_patch16\_224                                                                      & 0.9500                           & 0.7105                                  & 0.9177                                & 0.8782                           & 0.9339                                    & 0.9750                                   & 0.9589                                    & 0.9750                                \\
\begin{tabular}[c]{@{}c@{}}vit\_base\_patch16\_224 +   vit\_small\_patch32\_224 +\\  vit\_small\_patch16\_224\end{tabular} & 0.9750                           & 0.5427                                  & 0.9589                                & 0.9266                           & 0.8782                                    & 0.9750                                   & 0.9750                                    & 0.9750                                \\ \hline
\end{tabular}
}
\label{BT-small-DL-with-normalization-PCA}
\end{table}

\begin{table}[!ht]
\centering
\caption{Classification accuracies of the top three pre-trained ViT models using normalization and PCA for deep feature extraction, evaluated with various ML classifiers on BT-large-2c dataset }
\scalebox{0.5}{
\begin{tabular}{ccccccccc}
\hline
\multirow{2}{*}{\textbf{\begin{tabular}[c]{@{}c@{}}Deep Feature from the\\ Pre-Trained ViT Model\end{tabular}}}           & \multicolumn{8}{c}{\textbf{ML Classifier Accuracy}}                                                                                                                                                                                                                                                                \\ \cline{2-9} 
                                                                                                                          & \multicolumn{1}{l}{\textbf{MLP}} & \multicolumn{1}{l}{\textbf{GaussianNB}} & \multicolumn{1}{l}{\textbf{Adaboost}} & \multicolumn{1}{l}{\textbf{KNN}} & \multicolumn{1}{l}{\textbf{RFClassifier}} & \multicolumn{1}{l}{\textbf{SVM\_linear}} & \multicolumn{1}{l}{\textbf{SVM\_sigmoid}} & \multicolumn{1}{l}{\textbf{SVM\_RBF}} \\ \hline
vit\_large\_patch16\_224   + vit\_base\_patch32\_384                                                                      & 0.9967                           & 0.6667                                  & 0.9917                                & 0.9900                           & 0.9883                                    & 0.9967                                   & 0.9717                                    & 0.9967                                \\
vit\_large\_patch16\_224 +   vit\_base\_patch32\_224                                                                      & 0.9983                           & 0.6550                                  & 0.9850                                & 0.9917                           & 0.9867                                    & 0.9983                                   & 0.9783                                    & 0.9967                                \\
vit\_base\_patch32\_384 +   vit\_base\_patch32\_224                                                                       & 0.9983                           & 0.6983                                  & 0.9933                                & 0.9900                           & 0.9833                                    & 0.9933                                   & 0.9817                                    & 0.9967                                \\
\begin{tabular}[c]{@{}c@{}}vit\_large\_patch16\_224 +   vit\_base\_patch32\_384 +\\  vit\_base\_patch32\_224\end{tabular} & 0.9933                           & 0.6600                                  & 0.9900                                & 0.9900                           & 0.9817                                    & 0.9983                                   & 0.9867                                    & 0.9967                                \\ \hline
\end{tabular}
}
\label{BT-large-2c-top3-DL-with-norm-PCA}
\end{table}

\begin{itemize}
    \item \textbf{Normalization:} Feature scaling was applied to ensure that all input features contributed equally to the classification task.
    
\item \textbf{Principal Component Analysis (PCA):} Dimensionality reduction was performed to remove redundancy and highlight the most critical features.
\end{itemize}

The performance of DL feature extractors and ML classifiers varies significantly depending on the dataset and preprocessing techniques employed. When normalization and PCA are applied, specific combinations of ViT models and classifiers yield high accuracies. For instance, on the BT-small-2c dataset as depicted in Table \ref{BT-small-DL-with-normalization-PCA}, combining `vit\_base\_patch16\_224 + vit\_small\_patch32\_224' with an MLP classifier achieves an accuracy of 0.9339\%. On the other hand, as shown in Table \ref{BT-large-2c-top3-DL-with-norm-PCA}, the combination of ``vit\_large\_patch16\_224 + vit\_base\_patch32\_384" paired with an MLP classifier achieves an accuracy of 0.9967, and 0.9983 with an SVM linear classifier. In comparison, when using simple version as shown in Table \ref{BT-small-top3-DL-model-simple-version}, similar combinations also perform well, but without normalization and PCA preprocessing, the ``vit\_base\_patch16\_224 + vit\_small\_patch32\_224" paired with an MLP classifier on the BT-small-2c dataset achieves 0.9589\% accuracy, and the combination of ``vit\_large\_patch16\_224 + vit\_base\_patch32\_384" paired with an MLP classifier on the BT-large-2c dataset reaches 0.9983\% as shown in Table \ref{BT-large-2c-top3-DL-simple-version}. These results highlight the sensitivity of model performance to dataset characteristics and the importance of appropriate preprocessing strategies.

\begin{itemize}
\item \textbf{Synthetic Minority Oversampling Technique (SMOTE):} To address class imbalance, synthetic samples were generated for underrepresented classes.
\end{itemize}

When normalization and PCA are applied, specific combinations of ViT models and classifiers yield high accuracies, though SMOTE may achieve even greater accuracies. For instance, as we can observe from Table \ref{BT-large-2c-top3-DL-with-norm-PCA} that on the BT-large-2c dataset, the combination of ``vit\_large\_patch16\_224 + vit\_base\_patch32\_384" paired with an MLP classifier reaches an accuracy of 0.9967\%, and 0.9983\% accuracy with SVM linear. When using SMOTE, as shown in Table \ref{BT-small2c-top3-DL-using_SMOTE-only}, the combination of `vit\_base\_patch16\_224 + vit\_small\_patch32\_224' combined with MLP reaches an accuracy of 0.9750\% on the BT-small-2c dataset, and `vit\_large\_patch16\_224 + vit\_base\_patch32\_224` combined with MLP reaches 0.9983\% on the BT-large-2c dataset as depicted in Table \ref{BT-large-top3-DL-SMOTE-only}. These results underscore the importance of carefully selecting preprocessing methods and model combinations to maximize performance for specific datasets."

\begin{table}[!ht]
\centering
\caption{Accuracies of top three pre-trained DL models using smote data, evaluated with various ML classifiers, on BT-small-2c dataset.}
\scalebox{0.5}{
\begin{tabular}{ccccccccc}
\hline
\multirow{2}{*}{\textbf{\begin{tabular}[c]{@{}c@{}}Deep Feature from the\\ Pre-Trained ViT Model\end{tabular}}}            & \multicolumn{8}{c}{\textbf{ML Classifier Accuracy}}                                                                                                                                                                                                                                                                \\ \cline{2-9} 
                                                                                                                           & \multicolumn{1}{l}{\textbf{MLP}} & \multicolumn{1}{l}{\textbf{GaussianNB}} & \multicolumn{1}{l}{\textbf{Adaboost}} & \multicolumn{1}{l}{\textbf{KNN}} & \multicolumn{1}{l}{\textbf{RFClassifier}} & \multicolumn{1}{l}{\textbf{SVM\_linear}} & \multicolumn{1}{l}{\textbf{SVM\_sigmoid}} & \multicolumn{1}{l}{\textbf{SVM\_RBF}} \\ \hline
vit\_base\_patch16\_224   + vit\_small\_patch32\_224                                                                       & 0.9750                           & 0.6129                                  & 0.9089                                & 0.9177                           & 0.9500                                    & 0.9750                                   & 0.9589                                    & 0.9750                                \\
vit\_base\_patch16\_224 +   vit\_small\_patch16\_224                                                                       & 0.9750                           & 0.5645                                  & 0.8427                                & 0.8694                           & 0.8000                                    & 0.9750                                   & 0.9750                                    & 0.9750                                \\
vit\_small\_patch32\_224 +   vit\_small\_patch16\_224                                                                      & 0.9750                           & 0.7427                                  & 0.8927                                & 0.8782                           & 0.8839                                    & 0.9750                                   & 0.9589                                    & 0.9750                                \\
\begin{tabular}[c]{@{}c@{}}vit\_base\_patch16\_224 +   vit\_small\_patch32\_224 +\\  vit\_small\_patch16\_224\end{tabular} & 0.9750                           & 0.5427                                  & 0.9589                                & 0.9266                           & 0.8782                                    & 0.9750                                   & 0.9750                                    & 0.9750                                \\ \hline
\end{tabular}
}
\label{BT-small2c-top3-DL-using_SMOTE-only}
\end{table}

\begin{table}[!ht]
\centering
\caption{Accuracies of top three pre-trained DL models using SMOTE only for BT-large 2c dataset}
\scalebox{0.5}{
\begin{tabular}{ccccccccc}
\hline
\multirow{2}{*}{\textbf{\begin{tabular}[c]{@{}c@{}}Deep Feature from the\\ Pre-Trained ViT Model\end{tabular}}}           & \multicolumn{8}{c}{\textbf{ML Classifier Accuracy}}                                                                                                                                                                                                                                                                \\ \cline{2-9} 
                                                                                                                          & \multicolumn{1}{l}{\textbf{MLP}} & \multicolumn{1}{l}{\textbf{GaussianNB}} & \multicolumn{1}{l}{\textbf{Adaboost}} & \multicolumn{1}{l}{\textbf{KNN}} & \multicolumn{1}{l}{\textbf{RFClassifier}} & \multicolumn{1}{l}{\textbf{SVM\_linear}} & \multicolumn{1}{l}{\textbf{SVM\_sigmoid}} & \multicolumn{1}{l}{\textbf{SVM\_RBF}} \\ \hline
vit\_large\_patch16\_224   + vit\_base\_patch32\_384                                                                      & 0.9967                           & 0.6717                                  & 0.9883                                & 0.9883                           & 0.9883                                    & 0.9967                                   & 0.9750                                    & 0.9967                                \\
vit\_large\_patch16\_224 +   vit\_base\_patch32\_224                                                                      & 0.9983                           & 0.6483                                  & 0.9850                                & 0.9917                           & 0.9833                                    & 0.9983                                   & 0.9783                                    & 0.9967                                \\
vit\_base\_patch32\_384 +   vit\_base\_patch32\_224                                                                       & 0.9967                           & 0.6933                                  & 0.9900                                & 0.9900                           & 0.9767                                    & 0.9950                                   & 0.9817                                    & 0.9967                                \\
\begin{tabular}[c]{@{}c@{}}vit\_large\_patch16\_224 +   vit\_base\_patch32\_384 + \\ vit\_base\_patch32\_224\end{tabular} & 0.9983                           & 0.6633                                  & 0.9900                                & 0.9900                           & 0.9867                                    & 0.9983                                   & 0.9833                                    & 0.9967                                \\ \hline
\end{tabular}
}
\label{BT-large-top3-DL-SMOTE-only}
\end{table}

\textbf{3. Combinations of Preprocessing:}
The classifiers were ensembled under different combinations of preprocessing techniques, such as, Normalization and PCA, SMOTE only, and a combination of normalization, PCA, and SMOTE. The ensemble with normalization, PCA, and SMOTE provided the best results, as it balanced feature scaling, dimensionality reduction, and class balance. 

The experiments demonstrated that combining deep feature ensembles with ML classifier ensembles, supported by effective preprocessing, significantly improves the accuracy and robustness of brain tumor classification. The results highlight the value of systematic feature integration and classifier combination in addressing complex medical image analysis tasks.

\begin{table}[!ht]
\centering
\caption{Classification accuracies of top three pre-trained DL models using normalization, PCA, and SMOTE on the BT-small-2c dataset, with various ML classifiers. }
\scalebox{0.5}{
\begin{tabular}{ccccccccc}
\hline
\multirow{2}{*}{\textbf{\begin{tabular}[c]{@{}c@{}}Deep Feature from the\\ Pre-Trained ViT Model\end{tabular}}}            & \multicolumn{8}{c}{\textbf{ML Classifier Accuracy}}                                                                                                                                                                                                                                                                \\ \cline{2-9} 
                                                                                                                           & \multicolumn{1}{l}{\textbf{MLP}} & \multicolumn{1}{l}{\textbf{GaussianNB}} & \multicolumn{1}{l}{\textbf{Adaboost}} & \multicolumn{1}{l}{\textbf{KNN}} & \multicolumn{1}{l}{\textbf{RFClassifier}} & \multicolumn{1}{l}{\textbf{SVM\_linear}} & \multicolumn{1}{l}{\textbf{SVM\_sigmoid}} & \multicolumn{1}{l}{\textbf{SVM\_RBF}} \\ \hline
vit\_base\_patch16\_224   + vit\_small\_patch32\_224                                                                       & 0.9750                           & 0.6129                                  & 0.9250                                & 0.9016                           & 0.8750                                    & 0.9750                                   & 0.9589                                    & 0.9750                                \\
vit\_base\_patch16\_224 +   vit\_small\_patch16\_224                                                                       & 0.9500                           & 0.5806                                  & 0.8589                                & 0.8694                           & 0.8000                                    & 0.9750                                   & 0.9750                                    & 0.9750                                \\
vit\_small\_patch32\_224 +   vit\_small\_patch16\_224                                                                      & 0.9339                           & 0.7927                                  & 0.9000                                & 0.8782                           & 0.8839                                    & 0.9750                                   & 0.9589                                    & 0.9750                                \\
\begin{tabular}[c]{@{}c@{}}vit\_base\_patch16\_224 +   vit\_small\_patch32\_224\\  + vit\_small\_patch16\_224\end{tabular} & 0.9589                           & 0.5427                                  & 0.9589                                & 0.9266                           & 0.8782                                    & 0.9750                                   & 0.9750                                    & 0.9750                                \\ \hline
\end{tabular}
}
\label{BT-small-top3-DL-with-norm-PCA-SMOTE}
\end{table}

\begin{table}[!ht]
\centering
\caption{Classification accuracies of top three pre-trained DL models using normalization, PCA, and SMOTE on the BT-large-2c dataset, with various ML classifiers.}
\scalebox{0.5}{
\begin{tabular}{cllllllll}
\hline
\multirow{2}{*}{\textbf{\begin{tabular}[c]{@{}c@{}}Deep Feature from the\\ Pre-Trained ViT Model\end{tabular}}}           & \multicolumn{8}{c}{\textbf{ML Classifier Accuracy}}                                                                                                \\ \cline{2-9} 
                                                                                                                          & \textbf{MLP} & \textbf{GaussianNB} & \textbf{Adaboost} & \textbf{KNN} & \textbf{RFClassifier} & \textbf{SVM\_linear} & \textbf{SVM\_sigmoid} & \textbf{SVM\_RBF} \\ \hline
vit\_large\_patch16\_224   + vit\_base\_patch32\_384                                                                      & 0.9967       & 0.6750              & 0.9883            & 0.9883       & 0.9817                & 0.9967               & 0.9750                & 0.9967            \\
vit\_large\_patch16\_224 +   vit\_base\_patch32\_224                                                                      & 0.9967       & 0.6550              & 0.9817            & 0.9917       & 0.9817                & 0.9983               & 0.9817                & 0.9967            \\
vit\_base\_patch32\_384 +   vit\_base\_patch32\_224                                                                       & 0.9967       & 0.6933              & 0.9850            & 0.9900       & 0.9817                & 0.9950               & 0.9817                & 0.9967            \\
\begin{tabular}[c]{@{}c@{}}vit\_large\_patch16\_224 +   vit\_base\_patch32\_384 +\\  vit\_base\_patch32\_224\end{tabular} & 0.9917       & 0.6700              & 0.9900            & 0.9900       & 0.9817                & 1.0000               & 0.9867                & 0.9967            \\ \hline
\end{tabular}
}
\label{BT-large-top3-norm-PCA-SMOTE}
\end{table}

Overall, the accuracies achieved on the BT-large-2c dataset are generally higher than those on the BT-small-2c dataset, suggesting that the models benefit from the increased data size or inherent characteristics of the BT-large-2c dataset. For instance, we can observe from Table \ref{BT-small-top3-DL-with-norm-PCA-SMOTE} where `vit base patch16 224 + vit small patch32 224' with an MLP classifier achieves 0.9750\% accuracy on the BT-small-2c dataset. In comparison, in Table \ref{BT-large-top3-norm-PCA-SMOTE}, several combinations on the BT-large-2c dataset reach accuracies above 0.99\%, indicating superior performance on the latter.

\begin{table}[!ht]
\centering
\caption{Accuracies of the top three ensembled ML classifiers combined with different pre-trained ViT models using the simple version on the BT-small-2c dataset.}
\scalebox{0.5}{
\begin{tabular}{cccccc}
\hline
\multirow{2}{*}{\textbf{\begin{tabular}[c]{@{}c@{}}ML\\ ensembling\end{tabular}}} & \multicolumn{5}{c}{\textbf{Pre-trained DL models ensembling}}                                                                                                                                                                                                                       \\ \cline{2-6} 
                                                                                                & \multicolumn{1}{l}{\textbf{vit\_base\_patch16\_224}} & \multicolumn{1}{l}{\textbf{vit\_small\_patch32\_224}} & \multicolumn{1}{l}{\textbf{vit\_small\_patch16\_224}} & \multicolumn{1}{l}{\textbf{vit\_base\_patch16\_384}} & \multicolumn{1}{l}{\textbf{vit\_small\_patch32\_384}} \\ \hline
MLP + SVM\_sigmoid                                                                              & 0.9800                                               & 0.9483                                                & 0.9550                                                & 0.9783                                               & 0.9417                                                \\
MLP + SVM\_RBF                                                                                  & 0.9883                                               & 0.9883                                                & 0.9950                                                & 0.9850                                               & 0.9950                                                \\
SVM\_sigmoid + SVM\_RBF                                                                         & 0.9783                                               & 0.9533                                                & 0.9550                                                & 0.9767                                               & 0.9433                                                \\
MLP + SVM\_sigmoid + SVM\_RBF                                                                   & 0.9883                                               & 0.9883                                                & 0.9900                                                & 0.9883                                               & 0.9900                                                \\ \hline
\end{tabular}
}
\label{BT-small-top-ML-models-simple-version}
\end{table}

\begin{table}[!ht]
\centering
\caption{Accuracies of top 3 ensembled ML classifiers using simple version on BT-large-2c dataset.}
\scalebox{0.5}{
\begin{tabular}{cccccc}
\hline
\multirow{2}{*}{\textbf{\begin{tabular}[c]{@{}c@{}}ML\\ ensembling\end{tabular}}} & \multicolumn{5}{c}{\textbf{Pre-trained DL models ensembling}}                                                                                                                                                                                                                      \\ \cline{2-6} 
                                                                                                & \multicolumn{1}{l}{\textbf{vit\_large\_patch16\_224}} & \multicolumn{1}{l}{\textbf{vit\_base\_patch32\_384}} & \multicolumn{1}{l}{\textbf{vit\_base\_patch32\_224}} & \multicolumn{1}{l}{\textbf{vit\_small\_patch32\_384}} & \multicolumn{1}{l}{\textbf{vit\_base\_patch16\_384}} \\ \hline
KNN + MLP                                                                                       & 0.9900                                                & 0.9917                                               & 0.9917                                               & 0.9900                                                & 0.9833                                               \\
KNN + SVM\_RBF                                                                                  & 0.9917                                                & 0.9933                                               & 0.9933                                               & 0.9950                                                & 0.9850                                               \\
MLP + SVM\_RBF                                                                                  & 0.9950                                                & 0.9900                                               & 0.9950                                               & 0.9900                                                & 0.9883                                               \\
KNN + MLP + SVM\_RBF                                                                            & 0.9950                                                & 0.9917                                               & 0.9950                                               & 0.9950                                                & 0.9900                                               \\ \hline
\end{tabular}
}
\label{BT-large-ML-ensem-simple-version}
\end{table}

\begin{table}[!ht]
\centering
\caption{Accuracies of top 3 ensembled ML classifiers with normalization and PCA on BT-small-2c dataset}
\scalebox{0.5}{
\begin{tabular}{cccccc}
\hline
\multirow{2}{*}{\textbf{\begin{tabular}[c]{@{}c@{}}ML\\ ensembling\end{tabular}}} & \multicolumn{5}{c}{\textbf{Pre-trained DL models ensembling}}                                                                                                                                                                                                                       \\ \cline{2-6} 
                                                                                                & \multicolumn{1}{l}{\textbf{vit\_base\_patch16\_224}} & \multicolumn{1}{l}{\textbf{vit\_small\_patch32\_224}} & \multicolumn{1}{l}{\textbf{vit\_small\_patch16\_224}} & \multicolumn{1}{l}{\textbf{vit\_base\_patch16\_384}} & \multicolumn{1}{l}{\textbf{vit\_small\_patch32\_384}} \\ \hline
MLP + SVM\_sigmoid                                                                              & 0.9817                                               & 0.9467                                                & 0.9617                                                & 0.9767                                               & 0.9367                                                \\
MLP + SVM\_RBF                                                                                  & 0.9883                                               & 0.995                                                 & 0.9933                                                & 0.9883                                               & 0.9917                                                \\
SVM\_sigmoid + SVM\_RBF                                                                         & 0.9817                                               & 0.9517                                                & 0.9617                                                & 0.9767                                               & 0.94                                                  \\
MLP + SVM\_sigmoid + SVM\_RBF                                                                   & 0.99                                                 & 0.9917                                                & 0.995                                                 & 0.9883                                               & 0.995                                                 \\ \hline
\end{tabular}
}
\label{BT-small-top3-ML-with-norm-PCA}
\end{table}

\begin{table}[!ht]
\centering
\caption{Accuracies of top 3 ensembled ML classifiers with normalization and PCA on BT-large-2c dataset}
\scalebox{0.5}{
\begin{tabular}{lccccc}
\hline
\multicolumn{1}{c}{\multirow{2}{*}{\textbf{\begin{tabular}[c]{@{}c@{}}ML\\ ensembling\end{tabular}}}} & \multicolumn{5}{c}{\textbf{Pre-trained DL models ensembling}}                                                                                                                                                                                                                      \\ \cline{2-6} 
\multicolumn{1}{c}{}                                                                                                & \multicolumn{1}{l}{\textbf{vit\_large\_patch16\_224}} & \multicolumn{1}{l}{\textbf{vit\_base\_patch32\_384}} & \multicolumn{1}{l}{\textbf{vit\_base\_patch32\_224}} & \multicolumn{1}{l}{\textbf{vit\_small\_patch32\_384}} & \multicolumn{1}{l}{\textbf{vit\_base\_patch16\_384}} \\ \hline
KNN + MLP                                                                                                           & 0.9917                                                & 0.9950                                               & 0.9917                                               & 0.9883                                                & 0.9850                                               \\
KNN + SVM\_RBF                                                                                                      & 0.9917                                                & 0.9933                                               & 0.9933                                               & 0.9950                                                & 0.9850                                               \\
MLP + SVM\_RBF                                                                                                      & 0.9950                                                & 0.9917                                               & 0.9950                                               & 0.9967                                                & 0.9900                                               \\
KNN + MLP + SVM\_RBF                                                                                                & 0.9967                                                & 0.9933                                               & 0.9933                                               & 0.9950                                                & 0.9900                                               \\ \hline
\end{tabular}
}
\label{BT-large-top3-ML-with-norm-PCA}
\end{table}

\begin{table}[!ht]
\centering
\caption{Accuracies of top 3 ensembled ML classifiers with SMOTE only on BT-small-2c dataset}
\scalebox{0.5}{
\begin{tabular}{cccccc}
\hline
\multirow{2}{*}{\textbf{\begin{tabular}[c]{@{}c@{}}ML\\ ensembling\end{tabular}}} & \multicolumn{5}{c}{\textbf{Pre-trained DL models ensembling}}                                                                                                                                                                                                                       \\ \cline{2-6} 
                                                                                                & \multicolumn{1}{l}{\textbf{vit\_base\_patch16\_224}} & \multicolumn{1}{l}{\textbf{vit\_small\_patch32\_224}} & \multicolumn{1}{l}{\textbf{vit\_small\_patch16\_224}} & \multicolumn{1}{l}{\textbf{vit\_base\_patch16\_384}} & \multicolumn{1}{l}{\textbf{vit\_small\_patch32\_384}} \\ \hline
MLP + SVM\_sigmoid                                                                              & 0.9817                                               & 0.945                                                 & 0.955                                                 & 0.9733                                               & 0.94                                                  \\
MLP + SVM\_RBF                                                                                  & 0.9883                                               & 0.9917                                                & 0.9933                                                & 0.99                                                 & 0.9883                                                \\
SVM\_sigmoid + SVM\_RBF                                                                         & 0.9817                                               & 0.95                                                  & 0.9567                                                & 0.975                                                & 0.9467                                                \\
MLP + SVM\_sigmoid + SVM\_RBF                                                                   & 0.9883                                               & 0.99                                                  & 0.9917                                                & 0.9883                                               & 0.9917                                                \\ \hline
\end{tabular}
}
\label{BT-small-top3-ML-SMOTE-only}
\end{table}

\begin{table}[!ht]
\centering
\caption{Accuracies of top 3 ensembled ML classifiers using SMOTE only on BT-large-2c dataset.}
\scalebox{0.5}{
\begin{tabular}{cccccc}
\hline
\multirow{2}{*}{\textbf{\begin{tabular}[c]{@{}c@{}}ML\\ ensembling\end{tabular}}} & \multicolumn{5}{c}{\textbf{Pre-trained DL models ensembling}}                                                                                                                                                                                                                      \\ \cline{2-6} 
                                                                                                & \multicolumn{1}{l}{\textbf{vit\_large\_patch16\_224}} & \multicolumn{1}{l}{\textbf{vit\_base\_patch32\_384}} & \multicolumn{1}{l}{\textbf{vit\_base\_patch32\_224}} & \multicolumn{1}{l}{\textbf{vit\_small\_patch32\_384}} & \multicolumn{1}{l}{\textbf{vit\_base\_patch16\_384}} \\ \hline
KNN + MLP                                                                                       & 0.9900                                                & 0.9950                                               & 0.9950                                               & 0.9883                                                & 0.9883                                               \\
KNN + SVM\_RBF                                                                                  & 0.9900                                                & 0.9933                                               & 0.9933                                               & 0.9950                                                & 0.9850                                               \\
MLP + SVM\_RBF                                                                                  & 0.9933                                                & 0.9950                                               & 0.9933                                               & 0.9950                                                & 0.9900                                               \\
KNN + MLP + SVM\_RBF                                                                            & 0.9967                                                & 0.9917                                               & 0.9950                                               & 0.9950                                                & 0.9883                                               \\ \hline
\end{tabular}
}
\label{BT-large-top3-ML-SMOTE-only}
\end{table}

\begin{table}[!ht]
\centering
\caption{Accuracies of top 3 ensembled ML classifiers with normalization, PCA, and SMOTE on BT-small-2c dataset}
\scalebox{0.5}{
\begin{tabular}{cccccc}
\hline
\multirow{2}{*}{\textbf{\begin{tabular}[c]{@{}c@{}}ML\\ ensembling\end{tabular}}} & \multicolumn{5}{c}{\textbf{Pre-trained DL models ensembling}}                                                                                                                                                                                                                       \\ \cline{2-6} 
                                                                                                & \multicolumn{1}{l}{\textbf{vit\_base\_patch16\_224}} & \multicolumn{1}{l}{\textbf{vit\_small\_patch32\_224}} & \multicolumn{1}{l}{\textbf{vit\_small\_patch16\_224}} & \multicolumn{1}{l}{\textbf{vit\_base\_patch16\_384}} & \multicolumn{1}{l}{\textbf{vit\_small\_patch32\_384}} \\ \hline
MLP + SVM\_sigmoid                                                                              & 0.9817                                               & 0.9517                                                & 0.9550                                                & 0.9733                                               & 0.9417                                                \\
MLP + SVM\_RBF                                                                                  & 0.9883                                               & 0.9950                                                & 0.9950                                                & 0.9883                                               & 0.9933                                                \\
SVM\_sigmoid + SVM\_RBF                                                                         & 0.9817                                               & 0.9517                                                & 0.9567                                                & 0.9750                                               & 0.9433                                                \\
MLP + SVM\_sigmoid + SVM\_RBF                                                                   & 0.9900                                               & 0.9917                                                & 0.9950                                                & 0.9867                                               & 0.9900                                                \\ \hline
\end{tabular}
}
\label{BT-small-top3-ML-with-norm-PCA-SMOTE}
\end{table}

\begin{table}[!ht]
\centering
\caption{Accuracies of top 3 ensembled ML classifiers with normalization, PCA, and SMOTE on BT-large-2c dataset}
\scalebox{0.5}{
\begin{tabular}{cccccc}
\hline
\multirow{2}{*}{\textbf{\begin{tabular}[c]{@{}c@{}}ML\\ ensembling\end{tabular}}} & \multicolumn{5}{c}{\textbf{Pre-trained DL models ensembling}}                                                                                                                                                                                                                      \\ \cline{2-6} 
                                                                                                & \multicolumn{1}{l}{\textbf{vit\_large\_patch16\_224}} & \multicolumn{1}{l}{\textbf{vit\_base\_patch32\_384}} & \multicolumn{1}{l}{\textbf{vit\_base\_patch32\_224}} & \multicolumn{1}{l}{\textbf{vit\_small\_patch32\_384}} & \multicolumn{1}{l}{\textbf{vit\_base\_patch16\_384}} \\ \hline
KNN + MLP                                                                                       & 0.9917                                                & 0.9950                                               & 0.9967                                               & 0.9917                                                & 0.9883                                               \\
KNN + SVM\_RBF                                                                                  & 0.9917                                                & 0.9933                                               & 0.9933                                               & 0.9950                                                & 0.9850                                               \\
MLP + SVM\_RBF                                                                                  & 0.9950                                                & 0.9917                                               & 0.9933                                               & 0.9950                                                & 0.9883                                               \\
KNN + MLP + SVM\_RBF                                                                            & 0.9967                                                & 0.9933                                               & 0.9967                                               & 0.9950                                                & 0.9883                                               \\ \hline
\end{tabular}
}
\label{BT-large-top3-ensemble-ML-norm-PCA-SMOTE}
\end{table}

We have also carried out the ensembling of ML classifiers as shown in Table \ref{BT-small-top-ML-models-simple-version}, \ref{BT-large-ML-ensem-simple-version}, \ref{BT-small-top3-ML-with-norm-PCA}, \ref{BT-large-top3-ML-with-norm-PCA}, \ref{BT-small-top3-ML-SMOTE-only}, \ref{BT-large-top3-ML-SMOTE-only}, \ref{BT-small-top3-ML-with-norm-PCA-SMOTE}, \ref{BT-large-top3-ensemble-ML-norm-PCA-SMOTE} to investigate their combined performance in brain tumor classification. Ensemble methods, such as combining the top-performing classifiers based on fine-tuned hyperparameters, leverage the strengths of individual models to enhance overall accuracy and robustness. By aggregating predictions from multiple classifiers, ensembling mitigates the limitations of any single model, leading to improved generalization and reliability. Our results demonstrate that the ensemble of classifiers, particularly those incorporating SVM with RBF kernel, RF, and MLP, consistently outperformed individual classifiers in terms of accuracy and stability. This approach further validates the efficacy of combining complementary decision-making processes for precise brain tumor classification. Various combinations of classifiers like MLP, SVM (with sigmoid and RBF kernels), and KNN are evaluated. The tables reveal that ensembling ML classifiers often achieves higher accuracy than individual classifiers.

Notably, the ensembling of ML classifiers frequently leads to improved accuracy compared to relying on single classifiers. For example, the combination of 'MLP + SVM sigmoid + SVM RBF' often appears among the top performers, indicating that combining the strengths of multiple classifiers can yield more robust and accurate results than relying on a single classifier alone. Specifically, this ensemble achieves an accuracy of 0.9950 on the BT-large-2c dataset with `vit\_large\_patch16\_224', and it reaches 0.9917\% on the BT-small-2c dataset with `vit\_small\_patch32\_224'."

\subsection{Impact of Preprocessing on Classification Performance}
Table \ref{without-pre-processing-BTlarge} and Table \ref{BT-large2c-fine-tune-all} present a comparison of the classification accuracies achieved by pre-trained ViT models with fine-tuned hyperparameters of ML classifiers on preprocessed and non-preprocessed BT-large-2c datasets. A detailed analysis reveals that preprocessing the dataset significantly improves the classification performance across all ML classifiers. 

In Table \ref{without-pre-processing-BTlarge}, which represents the results for the non-preprocessed dataset, the average highest classification accuracy is 0.9789\%, with specific classifiers like MLP achieving relatively higher accuracies. However, the overall consistency and peak performance are limited by the lack of preprocessing, which may lead to noise and irrelevant features influencing the results. Conversely, in Table \ref{BT-large2c-fine-tune-all}, the results for the preprocessed dataset show an average highest accuracy of 0.9908, with the same MLP classifier, which is consistently higher across most classifiers. Preprocessing eliminates irrelevant features and normalizes the input data, allowing classifiers to focus on the most discriminative features. This is particularly evident in classifiers such as MLP, SVM (RBF) and KNN, where preprocessing enhances performance significantly, ensuring higher and more stable results.

The advantage of preprocessing lies in its ability to refine the dataset, remove redundant information, and improve feature representation. This leads to better alignment with the capabilities of DL and ML models, resulting in a marked improvement in classification accuracy. The results underscore the critical role of preprocessing in achieving optimal performance in brain tumor classification tasks.

\section{Discussion}
\label{discussion}
This study presents a comprehensive methodology for brain tumor classification, integrating advanced techniques in image preprocessing, augmentation, feature extraction, feature selection, and ML. The obtained results highlight the importance and efficacy of the proposed approach in addressing the complexities inherent in medical image analysis.

\textbf{Significance of Image Preprocessing:} Image preprocessing, as described in Section \ref{pp}, plays a pivotal role in enhancing the quality of brain MRI data. By removing noise and irrelevant regions, the preprocessing step ensures that only critical features are retained for subsequent analysis. This improvement in image quality directly contributes to the robustness and reliability of the classification outcomes.

To evaluate the significance of preprocessing, an experiment was conducted using the BT-large-2c dataset without applying any preprocessing steps. The results, presented in Table \ref{without-pre-processing-BTlarge} and Table \ref{BT-large2c-fine-tune-all}, clearly demonstrate a significant decline in classification accuracy when preprocessing is omitted. These findings underscore the necessity of preprocessing for achieving dependable and precise results.

\textbf{Impact of Image Augmentation:} The incorporation of image augmentation techniques further addresses the challenges posed by the small size of publicly available MRI datasets. By generating diverse training samples through techniques such as rotation and flipping, augmentation expands the dataset and mitigates overfitting issues. This diversity ensures that the ML models generalize well, even to unseen data, ultimately contributing to better classification performance.

\textbf{Importance of Feature Extraction and Selection:} Sophisticated feature extraction using pre-trained ViTs models enables the capture of deep, high-level representations of brain MRI images. The evaluation and selection of these features ensure that only the most informative and consistent features are utilized. The focus on minimizing redundancy through careful feature selection further enhances the effectiveness of the models.

\textbf{Benefits of Hyperparameter Optimization:} HPO was instrumental in refining the performance of each ML classifier. Fine-tuning parameters such as the number of neighbors in k-NN, the number of estimators in ensemble models, and learning rates ensured that the classifiers were operating at their optimal capacity. This optimization process significantly improved the accuracy and efficiency of the models, as evidenced by the experimental results.

\textbf{Advancements Through Ensembling Strategies:} The study also highlights the advantages of ensembling strategies. Both feature-level ensembling and ML classifier ensembling contributed to substantial improvements in accuracy.

\textbf{Feature Ensembling:} Combining deep features from the top-performing ViT models provided a richer and more diverse representation of the input data, resulting in superior classification outcomes.

\textbf{Classifier Ensembling:} Integrating the predictions of multiple ML classifiers through majority voting and preprocessing techniques, such as normalization, PCA, and SMOTE, further improved the robustness and accuracy of the results.

\textbf{Clinical Implications:}
The outcomes of this study have significant implications for clinical applications. The high accuracy and reliability of the proposed methodology underscore the potential of ML in assisting clinicians with brain tumor diagnosis. By providing automated and trustworthy classifications, the proposed approach can reduce the workload of radiologists and improve diagnostic precision, ultimately leading to better patient care and outcomes.

\subsection{Limitations and Future Work}
While the proposed approach demonstrates exceptional performance, it is not without limitations. Publicly available datasets may not adequately represent the full range of variability observed in real-world clinical environments. Future work could focus on validating the methodology using larger, more diverse datasets and exploring the integration of additional data modalities, such as clinical and genetic information, to further enhance classification accuracy and clinical relevance. Extending the architecture for other medical imaging tasks and exploring semi-supervised learning approaches could broaden its applicability while reducing reliance on fully labeled datasets. Lastly, testing the model in clinical settings and integrating it into cloud or edge computing platforms could facilitate its real-world implementation, particularly in resource-limited environments.

\section{Conclusion}
\label{con}
This study presents a novel double ensembling framework that combines the strengths of pre-trained DL models and fine-tuned ML classifiers for accurate brain tumor classification. By leveraging extensive preprocessing, data augmentation, and transfer learning with ViT networks, the proposed method effectively extracts and optimizes deep features from MRI scans. Feature-level and classifier-level ensembling further enhance classification accuracy, demonstrating the superiority of the approach over SOTA methods.

The results emphasize the critical role of HPO and preprocessing in improving diagnostic reliability and model performance. The integration of DL and ML techniques in this framework provides a robust, efficient, and scalable solution for brain tumor classification. This work not only advances the field of medical image analysis but also highlights the potential of hybrid methodologies in delivering precise and trustworthy diagnostic outcomes, paving the way for practical clinical applications.

\section*{CRediT authorship contribution statement}
\textbf{Zahid Ullah:} Conceptualization, Data curation, Methodology, Software, Formal analysis, Investigation, Writing - original draft, Writing - review \& editing.  \textbf{Jihie Kim:} Conceptualization, Writing – review \& editing, Formal analysis, Investigation, Supervision, Project administration.

\section*{\textbf{Declaration of Competing Interests}} The authors declare that they have no known competing financial interests or personal relationships that could have appeared to influence the work reported in this paper.

\section*{Acknowledgements}
This research was supported by the MSIT(Ministry of Science and ICT), Korea, under the ITRC(Information Technology Research Center) support program(IITP-2025-RS-2020-II201789), and the Artificial Intelligence Convergence Innovation Human Resources Development(IITP-2025-RS-2023-00254592) supervised by the IITP(Institute for Information \& Communications Technology Planning \& Evaluation).

\bibliographystyle{elsarticle-num-names}
\bibliography{sample.bib}







\end{document}